\documentclass[10pt]{article}
\title{ Long-Range Thermal 3D Perception in Low Contrast Environments}
\author{
  Filippov, Andrey\\
  \texttt{andrey@elphel.com} \\
  \normalsize{+1(801)783-5555x106}
  \and
  Filippova, Olga\\
  \texttt{olga@elphel.com} \\
  \normalsize{+1(801)783-5555x107}
}
\date{\normalsize{Elphel, Inc. 1455 West 2200 South \#205,
 West Valley City, UT 84119, USA}
}

\newif\ifprompts
\promptstrue

\newif\ifpromptsall
\promptsallfalse

\newif\ifproprietary
\proprietarytrue
\proprietaryfalse

\usepackage{multicol}

\newif\ifsigned
\signedfalse

\newif\ifbothproposals
\bothproposalstrue

\input{elphel-bib-glossary/helper_nsfformat.sty}
\usepackage[normalem]{ulem}
\usepackage{xfrac} 
\usepackage{nicefrac} 
\usepackage{enumitem}
\usepackage{gensymb}
\setitemize{noitemsep,topsep=0pt,parsep=0pt,partopsep=0pt}
\setenumerate{noitemsep,topsep=0pt,parsep=0pt,partopsep=0pt}
\usepackage{amsmath}
\usepackage{url, comment}
\usepackage[numbib,nottoc]{tocbibind}
\usepackage[breaklinks,pdfstartview=Fit,pagebackref=true]{hyperref}
\usepackage{wrapfig}
\usepackage{amssymb}
\usepackage{float}
\usepackage{times}

\usepackage{xspace}
\usepackage{fancyvrb}
\newsavebox{\verbbox}
\usepackage[explicit]{titlesec}
\usepackage[usenames,dvipsnames]{color}
\usepackage[table]{xcolor}
\usepackage{makecell}
\usepackage{longtable}
\usepackage{multirow}
\usepackage{pdfpages}
\usepackage{pbox}
\usepackage{titling} 

\usepackage{tocloft}
\usepackage{textgreek} 

\newcommand\milestonestable[3]{%
    \hline
      \multicolumn{1}{|c|}{\refstepcounter{MilestoneCounter}
      \label{milestone:#1}\textbf{M-\theMilestoneCounter}
         \expandafter\gdef\csname milestone#1description\endcsname{#3} 
         \expandafter\gdef\csname milestone#1due\endcsname{#2} 
        } &
      \begin{minipage}[t]{\linewidth}
      \textbf{#3}
      \end{minipage}
       & \multicolumn{2}{c|}{--} &
      \multicolumn{1}{c|}{\textbf{#2}} \\ 
}

\newcommand\milestonestablefoot[4]{%
    \hline
      \multicolumn{1}{|c|}{\refstepcounter{MilestoneCounter}
      \label{milestone:#1}\textbf{M-\theMilestoneCounter}
         \expandafter\gdef\csname milestone#1description\endcsname{#3} 
         \expandafter\gdef\csname milestone#1due\endcsname{#2} 
        } &
      \textbf{#3\footnote{#4}} & \multicolumn{2}{c|}{--} &
      \multicolumn{1}{c|}{\textbf{#2}} \\ 
}

\newcommand\scheduletable[2]{%
    \hline
      \multicolumn{1}{|c|}{
      M-\ref{milestone:#1} 
        } &
      \begin{minipage}[t]{\linewidth}
      \csname milestone#1description\endcsname
      \end{minipage}
       &
      #2 &
      \multicolumn{1}{c|}{\csname milestone#1due\endcsname} \\ 
}

\newcommand\scheduletablepay[5]{%
    \hline
      \multicolumn{1}{|c|}{
      M-\ref{milestone:#1} 
        } &
      \begin{minipage}[t]{\linewidth}
      \csname milestone#1description\endcsname
      \end{minipage}
      &
      \multicolumn{1}{c|}{\csname milestone#1due\endcsname} &  
      \multicolumn{1}{c|}{#2} &
      \multicolumn{1}{c|}{#3} &
      \multicolumn{1}{c|}{\$#4} &
      \multicolumn{1}{c|}{#5}
      \\
}

\newcommand\scheduletablenopay[4]{%
    \hline
      \multicolumn{1}{|c|}{
      M-\ref{milestone:#1} 
        } &
      \begin{minipage}[t]{\linewidth}
      \csname milestone#1description\endcsname
      \end{minipage}
      &
      \multicolumn{1}{c|}{\csname milestone#1due\endcsname} &  
      \multicolumn{1}{c|}{#2} &
      \multicolumn{1}{c|}{#3} &
      \multicolumn{1}{c|}{#4}
      \\
}

\usepackage{algorithm}
\usepackage[noend]{algpseudocode}

\usepackage{indentfirst}
\usepackage{graphicx}
\usepackage{grffile} 
\graphicspath{{Figures/}}
\DeclareGraphicsExtensions{.pdf,.jpeg,.png,.eps}

\usepackage{caption}
\usepackage{subcaption}

\usepackage{tikz}

\newcommand{\eat}[1]{}
\definecolor{Myorange}{cmyk}{0,0.42,1,0}

\newcommand{\eg}{\hbox{\em e.g.}}

\usepackage{needspace} 
\usepackage{pdflscape} 

\usepackage{pgfgantt}
\makeatletter
\@ifpackagelater{pgfgantt}{2018/01/01}{

}{%

}
\makeatother

\newganttlinktype{rdldr*}{%
  \draw [/pgfgantt/link]
    (\xLeft, \yUpper) --
    (\xLeft + \ganttvalueof{link bulge 1} * \ganttvalueof{x unit},
      \yUpper) --
    ($(\xLeft + \ganttvalueof{link bulge 1} * \ganttvalueof{x unit},
      \yUpper)!%
      \ganttvalueof{link mid}!%
      (\xLeft + \ganttvalueof{link bulge 1} * \ganttvalueof{x unit},
      \yLower)$) --
    ($(\xRight - \ganttvalueof{link bulge 2} * \ganttvalueof{x unit},
      \yUpper)!%
      \ganttvalueof{link mid}!%
      (\xRight - \ganttvalueof{link bulge 2} * \ganttvalueof{x unit},
      \yLower)$) --
    (\xRight - \ganttvalueof{link bulge 2} * \ganttvalueof{x unit},
      \yLower) --
    (\xRight, \yLower);%
}
\ganttset{
  link bulge 1/.link=/pgfgantt/link bulge,
  link bulge 2/.link=/pgfgantt/link bulge}

\definecolor{myblue}{rgb}{0.0, 0.3 0.5}
\hypersetup{
  colorlinks   = true, 
  urlcolor     = blue, 
  linkcolor    = myblue, 
  citecolor   =  red 
}

\usepackage{forarray} 

\newif\ifrev
\revtrue
\ifrev
  \newcommand{\PE}[1]{{\color{red} [PE: #1]}}
  \newcommand{\AL}[1]{{\color{green} [AL: #1]}}
  \newcommand{\AF}[1]{{\color{blue} [AF: #1]}}
  \newcommand{\OF}[1]{{\color{teal} [OF: #1]}}
  \newcommand{\OD}[1]{{\color{violet} [OD: #1]}}
\else
  \newcommand{\PE}[1]{}
  \newcommand{\AL}[1]{}
  \newcommand{\AF}[1]{}
  \newcommand{\OF}[1]{}
  \newcommand{\OD}[1]{}
\fi

\titleformat{\section}[hang]
{\normalfont\Large\bfseries}{}{0em}{\colorbox{black}{\parbox{\dimexpr\textwidth-2\fboxsep\relax}{\textcolor{white}{\thesection\quad#1}}}}{}

\titleformat{\subsection}[hang]
{\normalfont\large\bfseries}{\thesubsection}{0.5em}{#1}

\titleformat{\subsubsection} [runin] 
{\normalfont\normalsize\bfseries}{}{0em}{#1~~~}

\titlespacing*{\section}
{0pt}{0.7ex plus 1ex minus .2ex}{0.7ex plus .2ex}

\titlespacing*{\subsection}
{0pt}{0.4ex plus 0.5ex minus .2ex}{0.4ex plus .2ex}

\titlespacing*{\subsubsection}
{0pt}{0.3ex plus 0.3ex minus .1ex}{0.3ex plus .2ex}

\usepackage{textgreek} 
\usepackage{siunitx}
\usepackage[T1]{fontenc}
\DeclareSIUnit[number-unit-product = {}]{\inchQ}{\textquotedbl}
\DeclareSIUnit[number-unit-product = {\thinspace}]{\inch}{in}
\DeclareSIUnit[number-unit-product = {}]{\pixel}{pix}
\DeclareSIUnit[number-unit-product = {}]{\mph}{mph}
\DeclareSIUnit[number-unit-product = {}]{\electron}{$e^-$}

\usepackage[shortcuts,nopostdot,toc,nogroupskip,section,numberedsection=autolabel]{glossaries}
\setacronymstyle{long-short}

\makeglossaries 
\newacronym{uas}{UAS}{unpiloted aerial system}
\newacronym{uav}{UAV}{unpiloted aerial vehicle}
\newacronym{suas}{sUAS}{small unpiloted aerial system}
\newacronym{suass}{sUASS}{small unpiloted aerial system swarm}
\newacronym{uam}{UAM}{Urban Air Mobility}
\newacronym{aam}{AAM}{Advanced Air Mobility}

\newacronym
[description=counter-\gls{suass} system]
{csuass}{C-sUASS}
{Counter-sUASS}

\newacronym{lmfc}{LMFC}{Lockheed Martin Missile and Fire Control}
\newacronym{swap} {SWaP}{size, weight, and power}
\newacronym{swapc}{SWaP-C}{size, weight, power, and cost}

\newacronym
[description=light detection and ranging]
{lidar}{LiDAR}{light detection and ranging}

\newacronym
[description=high power continuous and pulsed laser and microwave weapon
systems] {dew}{DEW}{Directed Energy Weapons}

\newacronym{gps}{GPS}{Global Positioning System}
\newacronym{gnss}{GNSS}{Global Navigation Satellite System}

\newacronym{helmtt}{HELMTT}{High Energy Laser Mobile Test Truck}
\newacronym{mehel}{MEHEL}{Mobile Expeditionary High-Energy Laser}

\newacronym[description={short wave infrared (1.4-3.0~\textmu m)}]
{swir}{SWIR}{Short Wave Infrared}
\newacronym[description={middle wave infrared (3-8~\textmu m)}]
{mwir}{MWIR}{Middle Wave Infrared}
\newacronym[description={long wave infrared (8-15~\textmu m)}]
{lwir}{LWIR}{Long Wave Infrared}

\newacronym
[description={visible and near infrared, light bandwidth detected by ordinary
image sensors}]
{vnir}{VNIR}{Visible and Near Infrared}

\newacronym
[description={near infraredy}]
{nir}{NIR}{Near Infrared}

\newacronym
[description={visible range electro-optical imaging system}]
{eo}{EO}{electro-optical}

\newacronym
[description={electro-optical (visible range) and infrared imaging system}]
{eoir}{EO/IR}{Elecro-optical and Infrared}

\newacronym{adsb}{ADS-B}{{automatic dependent surveillance -- broadcast}}
\newacronym{spad}{SPAD}{single-photon avalanche diode}
\newacronym{fov}{FoV}{field of view}
\newacronym{hfov}{HFoV}{horizontal field of view}
\newacronym{vfov}{VFoV}{vertical field of view}
\newacronym{roi}{RoI}{region of interest}
\newacronym{tof}{ToF}{time-of-flight}
\newacronym{rf}{RF}{radio frequency}
\newacronym{dnn}{DNN}{deep neural network}
\newacronym{cnn}{CNN}{convolutional neural network}
\newacronym
[description={image sensor with the circuit board, lens and optional shutter}]
{sfe}{SFE}{sensor front end}

\newacronym
[description={central processing unit, main general purpose processor of a
system}] {cpu}{CPU}{central processing unit}

\newacronym
[description={graphic processing unit, used for  
massively parallel processing, including \gls{ml}}] {gpu}{GPU}{graphics
processing unit}

\newacronym{asic}{ASIC}{application-specific integrated circuit}

\newacronym{soc}{SoC}{system-on-chip}

\newacronym
[description={vision processing unit - \gls{asic} dedicated to image/video
processing, \gls{ml} acceleration}] {vpu}{VPU}{vision processing unit}

\newacronym
[description={Tile Processor - Elphel software/\gls{rtl} for 
transform-domain image processing}] {tp}{TP}{Tile Processor}

\newacronym{cots}{COTS}{commercial off-the-shelf}

\newacronym
[description={focal plane array - image sensor, including visible, infrared
and any other wavelength}] {fpa}{FPA}{focal plane array}

\newacronym{piv}{PIV}{Particle Image Velocimetry}
\newacronym{psf}{PSF}{point spread function}
\newacronym{cad}{CAD}{computer aided design}
\newacronym{sn}{S/N}{signal-to-noise}
\newacronym{snr}{SNR}{signal-to-noise ratio}

\newacronym
[description={Electronic Rolling Shutter - image sensor exposure method, where
pixels are exposed in line-scan order (opposite to snapshot shutter)}]
{ers}{ERS}{electronic rolling shutter}

\newacronym
[description={Frequency Domain, usually same as Transform Domain - data
transformed by Fourier or similar transform, enables efficient
implementation of convolution and correlation}]
{fd}{FD}{Frequency Domain}

\newacronym{fdshort}{FD}{frequency domain}

\newacronym{dct}{DCT}{discrete cosine transform}
\newacronym{dst}{DST}{discrete sine transform}
\newacronym
[description={Discrete Trigonometric Transform - common name for shared by 
\gls{dct} and \gls{dst}}]
{dtt}{DTT}{discrete trigonometric transform}

\newacronym
[description={Modified Complex Lapped Transform - invertible transform
to/from \gls{fd} based on \gls{dct} and \gls{dst} type~IV}] {mclt}{MCLT
}{modified complex lapped transform }

\newacronym{dofr}{DoF}{degrees of freedom}

\newacronym{tdac}{TDAC}{time-domain aliasing cancellation}

\newacronym{ssd}{SSD}{solid-state drive}

\newglossaryentry{jp4}{name={JP4}, description={image format
developed by Elphel for efficient compression of raw Bayer mosaic images; first
used for Google Books project}}

\newacronym{rmse}{RMSE}{root mean square error}

\newacronym
[description={register-transfer level - design abstraction level in
hardware description languages}] {rtl}{RTL}{register-transfer level}

\newacronym{floss}{FLOSS}{free, libre, and open-source software}
\newacronym{fpga}{FPGA}{field-programmable gate array}
\newacronym{pld}{PLD}{programmable logic devices}

\newacronym{ml}{ML}{machine learning}
\newacronym{lma}{LMA}{Levenberg-Marquardt algorithm}

\newglossaryentry{tpnet}{name={TPNET}, description={\gls{tp} with \gls{dnn}
for 3D perception system}}

\newacronym
[description={Google \gls{asic} for \gls{ml} acceleration}]
{tpu}{TPU}{tensor processing unit}
\newacronym{ai}{AI}{artificial intelligence}
\newacronym{ugv}{UGV}{unpiloted ground vehicle}
\newacronym{agv}{AGV}{autonomous ground vehicle}
\newacronym{slam}{SLAM}{simultaneous localization and mapping}

\newglossaryentry{sonar}{name={sonar}, description={ultrasound ranging  device,
originally acronym for SOund NAvigation Ranging}}

\newacronym{ar}{AR}{augmented reality}
\newacronym{pcb}{PCB}{printed circuit board}
\newacronym{ara}{ARA}{Applied Research Associates}
\newacronym{poc-corp}{POC}{Physical Optics Corporation}
\newacronym{imu}{IMU}{inertial measurement unit}
\newacronym{ims}{IMS}{inertial measurement system}
\newacronym{aas}{AAS}{autonomous amphibious system}
\newglossaryentry{lm-athena}{name={ATHENA}, description={Lockheed
Martin Advanced Test High Energy Asset}}
\newglossaryentry{ray-phaser}{name={PHASER}, description={Raytheon
anti-drone high-powered microwave emitter}}

\newglossaryentry{gnugpl}{name={GNU/GPL}, description={GNU General Public
License}}
\newglossaryentry{cernohl}{name={CERN OHL}, description={CERN (European
Organization for Nuclear Research) Open Hardware License}}
\newacronym{atak}{ATAK}{Android Team Awareness Kit}
\newacronym{afrl}{AFRL}{Air Force Research Laboratory}
\newacronym{anit}{ANIT}{alternative night/day imaging technologies}
\newacronym
[description={U.S. Army Futures Command Integrated Visual Augmentation
System}]{ivas}{IVAS}{Integrated Visual Augmentation System}
\newacronym{hud3}{HUD 3.0}{Heads Up Display 3.0}
\newacronym{fps}{FPS}{frames per second}
\newacronym{sbir-ca}{CA}{contract award}
\newacronym{rv}{RV}{reentry vehicle}
\newacronym{sbir}{SBIR}{Small Business Innovation Research}
\newacronym{em}{EM}{electromagnetic}
\newacronym{sfm}{SfM}{Structure-from-Motion}
\newacronym{hgv}{HGV}{Hypersonic Glide Vehicle}
\newacronym{ssl-mda}{SSL}{Space Sensor Layer}
\newacronym{hbtss}{HBTSS}{Hypersonic and Ballistic Tracking Space Sensor}
\newacronym{pleo}{P-LEO}{Proliferated Low Earth Orbit}
\newacronym{mre}{MRE}{mean reprojection error}
\newacronym{ti}{TI}{thermal imaging}
\newacronym
[description={differential rectification of the stereo images to
the common for the set distortion model}] {difrec}{DR}{differential
rectification}
\newacronym{rsd}{RSD}{relative standard deviation}
\newacronym{svd}{SVD}{singular value decomposition}
\newacronym{fc}{FC}{fully connected}
\newacronym{fg}{FG}{foreground}
\newacronym{bg}{BG}{background}
\newacronym{stp}{STP}{Soldier Touch Point}
\newacronym{tsoa}{TSOA}{Technical Support and Operational Analysis}
\newacronym{sgm}{SGM}{Semi-Global Matching}
\newacronym{asa}{ASA}{adaptive squad architecture}

\newacronym{vts}{VTS}{Virtual Tradespace}
\newacronym{ih}{IH}{Invisible Headlights}
\newacronym{gt}{GT}{ground truth}
\newacronym{netd}{NETD}{noise-equivalent temperature difference}
\newacronym{sar}{SAR}{synthetic-aperture radar}
\newacronym{ms}{MS}{multi-sensor}
\newacronym{mb}{MB}{motion blur}
\newacronym{gfi}{GFI}{Government-Furnished Information} 
\newacronym{gfe}{GFE}{Government-Furnished Equipment} 
\newacronym{ttp}{TTP}{targeting task performance, as used by NVESD}
\newacronym
[description={figure of merit for \gls{lwir} sensors, equal to
$\gls{netd}\times\tau$, measured in mK$\cdot$ms}]{fom}{FOM}{figure of merit}
\newacronym{mems}{MEMS}{micro-electromechanical system}
\newacronym{ekf}{EKF}{extended Kalman filter}
\newacronym{icp}{ICP}{iterative closest point}
\newacronym{pnt}{PNT}{position, navigation, and timing}
\newacronym{apnt}{A-PNT}{assured-position, navigation, and timing}
\newacronym{van}{VAN}{vision aided navigation}
\newacronym{ins}{INS}{inertial navigation system}
\newacronym{dtc}{DTC}{Detection, Tracking, and Classification}
\newacronym{soa}{SOA}{State of the Art} 
\newacronym{s-o-a}{SOA}{state-of-the-art}
\newacronym{nasa_ttt}{TTT}{NASA Transformational Tools and Technologies project}
\newacronym{nasa_nas}{NAS}{National Air Space}

\newacronym{adas}{ADAS}{advanced driver-assistance systems}
\newacronym{dof}{DOF}{depth of field}
\newacronym{envg}{ENVG}{Enhanced Night Vision Goggle}
\newacronym{envgb}{ENVG-B}{Enhanced Night Vision Goggle - Binocular}
\newacronym{taba}{TABA}{Technical and Business Assistance}
\newacronym{gan}{GAN}{Generative Adversarial Network}
\newacronym{gdsr}{GDSR}{Guided Depth Super-Resolution}

\iftrue
\newglossarystyle{longwide}{%
     {\begin{longtable}{lp{\glsdescwidth}}}%
     {\end{longtable}}%
}
\fi
\usepackage [square,sort,comma,numbers]{natbib}
\usepackage{fancyhdr}
\usepackage{nameref}
\renewcommand{\glossarysection}[2][]{}

\usepackage[english]{babel}
\usepackage{blindtext}

\usepackage{sf298}

\begin{document}
\glsunset{swap}
\glsunset{swapc}
\glsunset{cots}
\glsunset{lidar}
\glsunset{ai}
\glsunset{ml}
\glsunset{ar}
\glsunset{asic}
\glsunset{gps}

\maketitle

\begin{center}
\begin{center}
	\includegraphics[width=0.3\linewidth]{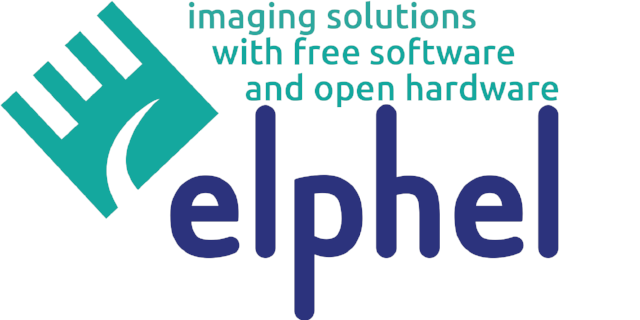}
	\end{center}
	
\large\textbf{NASA SBIR 2021-I Phase I Final Report}\\
\large\textbf{November 19, 2021}\\

\end{center}
\vspace{10pt}

\begin{abstract}This report discusses the results of Phase~I effort to
prove the feasibility of dramatic improvement of the microbolometer-based \gls{lwir}
detectors sensitivity, especially for the 3D measurements. The resulting low
\gls{swapc} thermal depth-sensing system will enable the situational awareness of
Autonomous Air Vehicles for \gls{aam}. It will provide robust 3D
information of the surrounding environment, including low-contrast static and
moving objects, at far distances in degraded visual conditions and \gls{gps}-denied areas.
Our multi-sensor 3D perception enabled by \gls{cots} uncooled thermal sensors
mitigates major weakness of \gls{lwir} sensors - low contrast by increasing the system
sensitivity over an order of magnitude.

There were no available thermal image sets suitable for evaluating this
technology, making datasets acquisition our first goal.
We discuss the design and construction of the prototype system with sixteen
\SI{640x512}{\pixel} \gls{lwir} detectors, camera calibration to subpixel resolution,
capture, and process synchronized image. The results show the $3.84\times$
contrast increase for intrascene-only data and an additional $5.5\times$ – with
the interscene accumulation, reaching system \gls{netd} of
\SI{1.9}{\milli\kelvin} with the \SI{40}{\milli\kelvin} sensors.
\end{abstract}

\newpage

\iftrue
\phantomsection 
\addcontentsline{toc}{section}{Contents}
\tableofcontents
\fi

\iftrue
\phantomsection 
\listoffigures
\fi
\newpage

\section{Project Summary}
\label{sec:proj_sum}

\noindent
\textbf{Firm: Elphel, Inc. }\\
\textbf{Project Title: Long Range Thermal 3D Perception in Low Contrast Environments}\\

\subsection{Identification and Significance of Innovation}
\label{subsection:sum_identification}

Long-range 3D perception is an essential capability of the intelligent vehicle
systems enabling situational awareness of the autonomous and piloted air
vehicles for NASA \gls{aam}. Current depth sensing and exterior
perception technologies are short-range automotive \glspl{lidar} and long-range radars
with limited transversal and temporal resolution, especially in the dense urban
environment, do not adequately respond to AAM requirements. High travel speeds
combined with the relatively low deceleration rates require long sensing
distances, and the range of the conventional LiDAR measurements is often
insufficient. We propose to provide future air vehicles with \gls{lwir} 3D perception
capability that enables vehicle awareness of the environment in degraded visual
conditions, in the absence of ambient illumination, robust against \gls{gps}
failure increasing the safety of \gls{nasa_nas}.
Thermal images can be used similarly to the visible range ones for 3D scene
reconstruction, but the challenges specific for \gls{lwir} - lower image resolution,
lower contrast of the textures, and high thermal inertia, so far prevented its
use for robust 3D.
We have invented a novel method for thermal imaging contrast improvement over an
order of magnitude, which will revolutionize thermal imaging and enable
long-range 3D perception with uncooled \gls{lwir} sensors.

\subsection{List of Technical Objectives}
\label{subsection:sum_objectives}
The main objectives of the Phase~I project are to demonstrate $10\times$ --
$20\times$ improvement of the microbolometer-based \gls{lwir} detectors'
system sensitivity and increase the ranging accuracy.

There were no available thermal image sets that we could use to evaluate the
proposed technology, making such datasets acquisition our first objective. Our
approach presents unique requirements for the captured data: a large number of
synchronized non-collinear images (16 + 4 apertures) and strict requirements for
the camera calibration (\SI{0.02}{\pixel} reprojection \gls{rmse}). To achieve
the objectives of Phase~I, we assembled an experimental dual-modal
(\gls{lwir}/\gls{eo}) prototype system shown in Figure~\ref{fig:prototype},
acquired the representative image data sets, and implemented image processing
algorithms to improve contrast with intrascene (using multiple simultaneous
sensor images) and interscene (consolidating data from multiple consecutive
scenes) methods.

We evaluated the contrast improvement as system \gls{netd} and compared it to
the traditional stereo configuration and methods.

We expected to demonstrate $4\times$ contrast improvement over the binocular
configuration for the full 16-sensor scenes in this test, corresponding to the
effective \gls{netd} of \SI{10}{\milli\kelvin}.  With the interscene
accumulation, we planned to achieve a system \gls{netd} of \SI{1}{\milli\kelvin}
to \SI{2}{\milli\kelvin}, depending on the number of accumulated scenes,
camera movement, and the \gls{lwir} sensor module internal properties.

The above method and camera configuration allow developing a scalable Tradespace
model, evaluating subsets of 2 (binocular), 4, 8, and all 16 sensors and
comparing the achieved RMSE over the full scene FoV.

\subsection{List of Technical Acomplishments}
\label{subsection:sum_tech_results}
\begin{enumerate}
  \item We have assembled the dual-modal (\gls{lwir}/\gls{eo}) prototype camera,
  shown in Figure~\ref{fig:prototype}, for the experimental datasets'
  acquisition.
  \item Performed photogrammetric camera calibration using our dedicated
  calibration setup with a large \SI{7x3}{\meter} \gls{lwir}/\gls{eo}
  calibration pattern.
  \item Acquired \gls{lwir}/\gls{eo} image data sets in various field conditions
  to collect representative data.
  \item Updated the software to accommodate processing from 16 \gls{lwir}
  simultaneously acquired images.
  \item Evaluated thermal contrast improvement of 3D thermal perception system
  over the existing methods.
  \item We have achieved the stated metrics of $3.84\times$ increasing the
  contrast gain with the intrascene method and $5.5\times$ additional contrast
  improvement with the interscene accumulation of measurements. Our results show
  \SI{10.5}{\milli\kelvin} for intrascene and \SI{1.9}{\milli\kelvin} with the
  interscene methods.
  \item Our results prove the feasibility of the proposed technology.
\end{enumerate}

\subsection{Name and Address of Principal Investigator}
\label{subsection:sum_PI}

Dr. Andrey Filippov

1455 West 2200 South \#205, West Valley City, UT 84119, USA

\subsection{Name and Address of Offeror}
\label{subsection:sum_offeror}

Elphel, Inc.

1455 West 2200 South \#205, West Valley City, UT 84119, USA

\newpage

\section{Abstract}
\label{sec:abstract}

\iftrue
\begin{wrapfigure}{L}{0.45\textwidth} 
\vspace{-10pt}
\centering \includegraphics[width=0.97\linewidth]{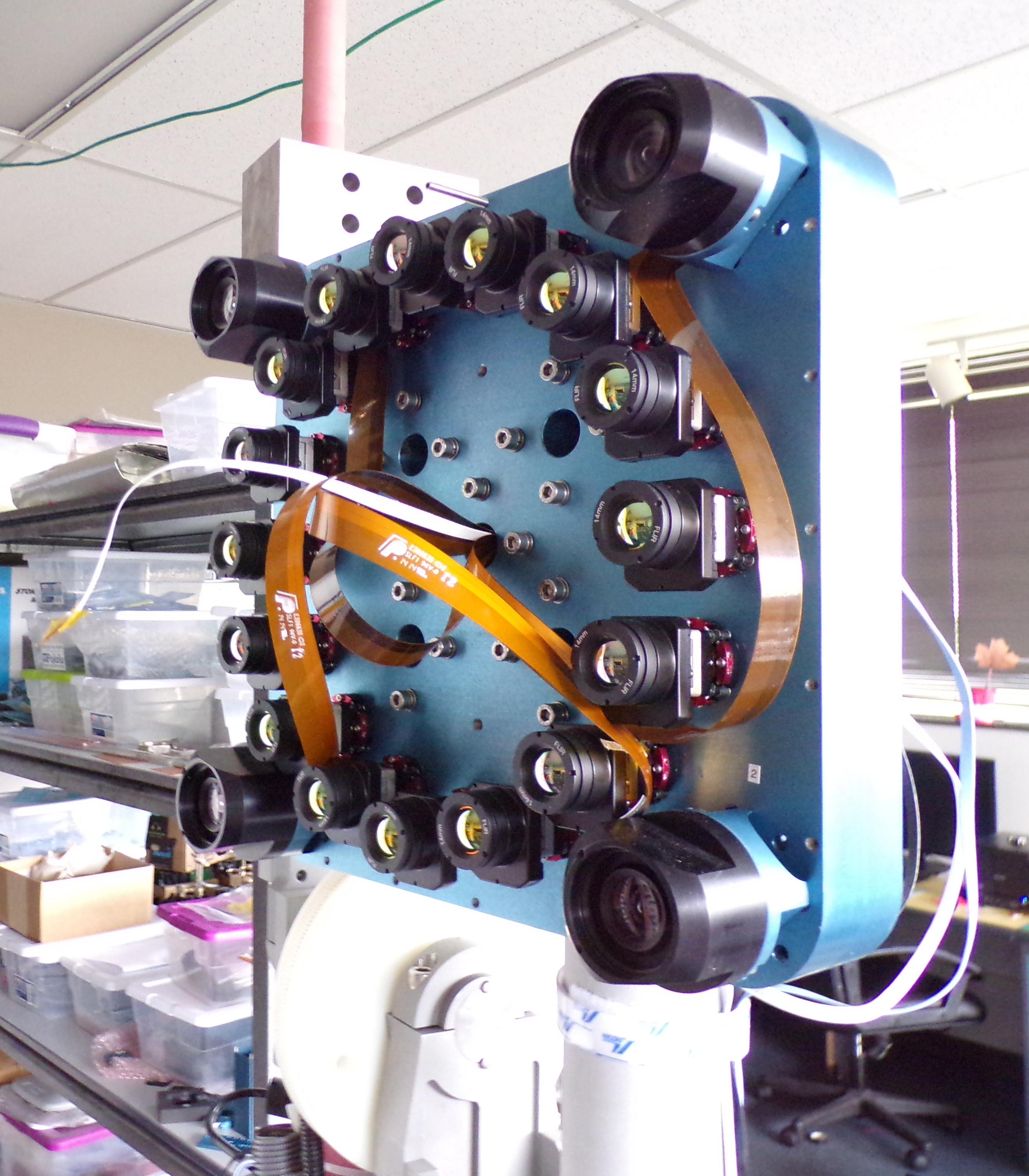}
\captionof{figure}{16 \gls{lwir} sensor + quad \gls{eo} prototype.}
\label{fig:prototype}
\vspace{-10pt}
\end{wrapfigure}
\fi
\Gls{lwir} cameras provide images regardless of the ambient
illumination; they tolerate fog, and the incoming car
headlights do not blind them. These features make \gls{lwir} cameras attractive
for autonomous navigation of air and ground vehicles. Thermal images can be used
similarly to the visible range ones for 3D scene reconstruction with passive (not
emitting IR radiation) methods, but the challenges specific for \gls{lwir} - lower image
resolution, lower contrast of the textures, and high thermal inertia, so far
prevented its use for robust 3D.

Our technology mitigates these limitations of the uncooled thermal sensors. The
novel sensor configuration, calibration methods, and image processing algorithms
improve the microbolometer-based 3D perception, surpassing the traditional
stereo build with the most advanced sensors, even those limited only by the
photon detectors' shot noise. The invented method of interscene accumulation of
the intrascene correlations will result in the system's \gls{netd} of
\SI{1}{\milli\kelvin} and better.

The novelty and uniqueness of the proposed work submitted in part as a
US patent application ``Method for the 3D Thermal Imaging Motion Blur
Mitigation and Contrast Enhancement'' is in the method of how we increase the
contrast of the thermal images and compensate for the thermal inertia of the
uncooled image sensors by use of massive noncollinear arrays of thermal sensors.
The conventional binocular stereo systems fail to fully use the image pair's
captured data as only the mismatch along the epipolar lines contributes to the
depth calculation. Our current research shows that even the quadocular stereo
camera reduces effective \gls{netd} twice (not as $\sqrt2$) compared to 
the binocular configuration, while the number of sensors is only two times
higher. The 16-sensor prototype will improve effective \gls{netd} four times
relative to a conventional binocular system, reaching \SI{10}{\milli\kelvin}.
Interscene accumulation of the 2D intrascene (single-shot) correlation output
improves sensitivity by another order of magnitude.

\section{Technical Objectives}
\label{sec:objectives}

The main research objective in Phase~I is to prove the feasibility of dramatic
(over an order of magnitude) improvement of the microbolometer-based \gls{lwir}
detectors sensitivity, especially for the 3D measurements. This perception
technology targets the critical for the \gls{aam} applications gap between
currently available short-range automotive \glspl{lidar} and long-range radars
with the limited transversal and temporal resolution, especially for the
cluttered urban environment.

\subsection{Approach}
\label{subsection:approach}
We will use a massive multi-aperture configuration of accurately calibrated uncooled LWIR
sensors to enable 3D perception in extremely low-contrast environments at long
range and high speed.

The proposed approach presents unique requirements for the captured data: a
large number of the synchronized noncollinear images (preferably 16 or more) and
strict requirements for the camera calibration (0.025pix reprojection
\gls{rmse}) make achieving the project goals with currently available datasets
impossible. We will use the data acquisition prototype designed and assembled as part of this research
(Figure~\ref{fig:prototype}) to demonstrate an intrascene system \gls{netd}
of \SI{10}{\milli\kelvin} and reach \SI{1}{\milli\kelvin} with the interscene
accumulation for the environment's low-contrast features.

The \gls{lwir} subsystem containing sixteen FLIR Boson \SI{640x512}{\pixel}
camera cores (Figure~\ref{fig:prototype}) will provide image data to demonstrate
system \gls{netd}\footnote{The sensitivity of the hypothetical sensors in a
binocular stereo configuration capable of ranging the same scene.}of
\SI{10}{\milli\kelvin} in the intrascene-only correlation mode. With the
interscene accumulation possible for terrain (static) features with the initial
software that is possible to upgrade for most moving objects, we expect to
achieve system \gls{netd} of \SIrange{1}{2}{\milli\kelvin} depending on the
number of accumulated scenes.

With the \SI{32}{\degree}~\gls{hfov} and \SI{220}{\milli\meter}
diameter of the lens centers circle, this small-size prototype will provide
dense depth maps with \SI{10}{\percent} accuracy at \SI{500}{\meter} range
using Phase~I's preliminary software and disparity resolution of
\SI{0.05}{\pixel}. In Phase~II, we expect to improve disparity resolution to
\SI{0.025}{\pixel} using a trained \gls{dnn}
and achieve \SI{5}{\percent} accuracy at
\SI{500}{\meter}.

The proposed effort aims to develop low SWaP-C day/night, all-weather
depth-sensing hardware that can detect objects at far distances, e.g.,
1500~ft to 3~nmi with passive \gls{lwir} sensors.

\needspace{15\baselineskip}

\section{Technical Activities}
\label{sec:tech-activities}

\subsection{Multi-Sensor Camera Assembly and Testing}
\label{sub:multi-sensor}
To achieve the objectives of Phase 1, we assembled an experimental
dual-modal (\gls{lwir}/\gls{eo}) prototype system shown in
Figure~\ref{fig:prototype}.
We designed this system to use 16 identical microbolometer-based FLIR Boson camera
cores (\SI{640x512}{\pixel}) evenly distributed over the circle of
\SI{110}{\milli\meter} radius and four \SI{5}{\mega|pixel} RGB sensors
positioned in the corners.

\iftrue
\begin{wrapfigure}{R}{0.42\textwidth} 
\vspace{-10pt}
\centering \includegraphics[width=0.97\linewidth]{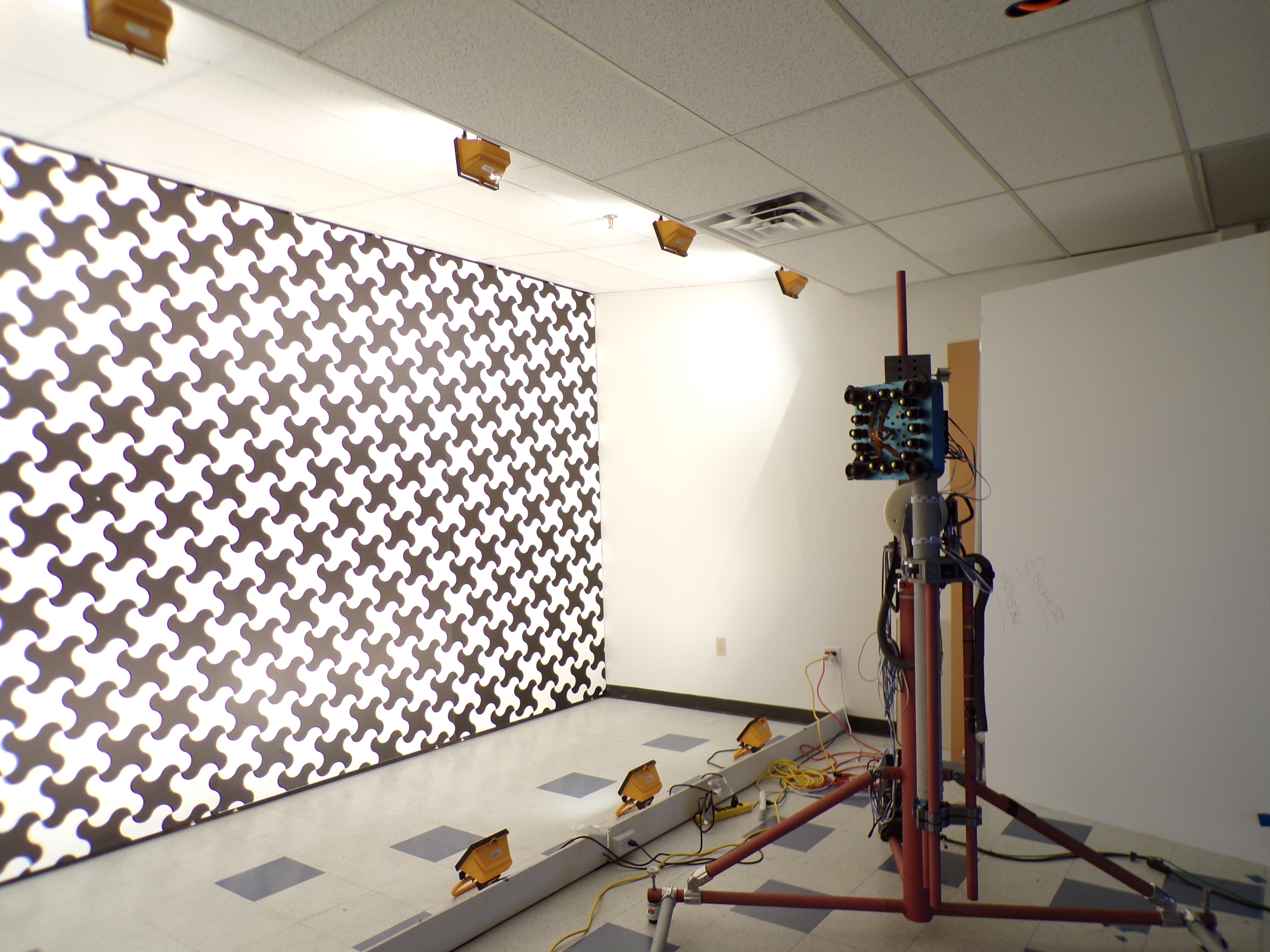}
\captionof{figure}{16 LWIR prototype camera calibration.}
\label{fig:16-lwir-calibration}
\vspace{-10pt}
\end{wrapfigure}
\fi

We have thoroughly tested these RGB sensors for long-range 3D sensing; they will
provide ground truth measurements for thermal sensors during daytime operation.
All the optical components are mounted on a \SI{50}{\milli\meter}-thick base
plate with an internal honeycomb structure to maintain the sensor modules'
mechanical stability within \SI{10}{\micro\radian} needed for
\SI{0.02}{\pixel} disparity resolution of the \SI{32}{\degree} \gls{hfov} \SI{640x512}{\pixel}
thermal sensors.
Each sensor module has a 3-axis attitude adjustment, enabling their alignment
during the factory calibration procedure. While the
\gls{difrec}~\cite{filippov2019method} method of image matching tolerates minor
orientation errors, it still requires the imagers' mechanical alignment. In
addition to the sensor components, the prototype system includes five Elphel
NC393 quad-port camera system modules to acquire and record image data to the
internal \gls{ssd} devices. One such module operates four RGB sensors, and
the other four control sixteen \gls{lwir} sensors (four each) and record
images to the \glspl{ssd}. Each sensor unit operates in the external trigger
mode synchronized by the master camera. \gls{lwir} modules run at
\SI{60}{\hertz}; the RGB ones capture 15 frames per second. The
synchronization network distributes shared \SI{1}{\micro\second}-accurate
timestamps for embedding into each image file metadata, simplifying scene
assembly from multiple storage devices.

\subsection{Camera Calibration}
\label{sub:calibration}

Photogrammetric camera calibration is a precondition to achieve an accurate
depth map in the stereo application, and most methods use dedicated calibration
setups to perform this task.

We perform camera calibration with a large LWIR/RGB calibration pattern and
goniometer for precise automatic rotation.  Our dual-modality \SI{7.0x3.0}{\meter}
calibration target consists of five separate panels fitted together.
We use twelve 500 W halogen floodlights to illuminate the pattern printed on a 5
mm foamcore board glued to a 0.8 mm aluminum sheet. The forced airflow cools its
back side.

The black-and-white pattern resembles the checkerboard one with each straight
edge replaced by a combination of two arcs – this makes spatial spectrum uniform
and facilitates point spread function (PSF) measurement.

We use an automated goniometer (Figure~\ref{fig:16-lwir-calibration}) to
capture many (typically hundreds) image sets so that in captured data, each area
of the camera sensor \gls{fov} captures most of the large pattern. This allows
accurate bundle adjustment to determine camera intrinsic and extrinsic parameters
simultaneously with measuring the target pattern itself.

\iftrue
\begin{wrapfigure}{L}{0.40\textwidth} 
\centering \includegraphics[width=0.97\linewidth]{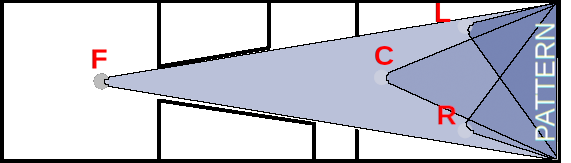}
\captionof{figure}{Camera positions during calibration: L- to the left of the
centerline, 4.5m from the target; R - to the right of the centerline, 4.5m from the
target; C - centerline, 7.4m from the target, F - centerline, 22 m from the
target.}
\label{fig:stations_plan}
\vspace{-10pt}
\end{wrapfigure}
\fi

It is not practical
to build an ``ideal `` pattern with the accuracy sufficient for deep subpixel
calibration, so instead, we measure the 3D location of each pattern grid node to
approximately \SI{0.1}{\milli\meter} accuracy. As a result of the calibration, we
measure:
\begin {enumerate}
  \item target pattern nodes coordinates;
  \item each camera module radial distortion model parameters;
  \item each camera module deviations from the radial distortion model;
  \item each camera module location and orientation in the camera coordinate
  system;
  \item space-variant \gls{psf} of each camera module for image aberration
  correction.
\end{enumerate}

\iftrue
\begin{wrapfigure}{R}{0.45\textwidth} 
\vspace{-10pt}
\centering \includegraphics[width=0.97\linewidth]{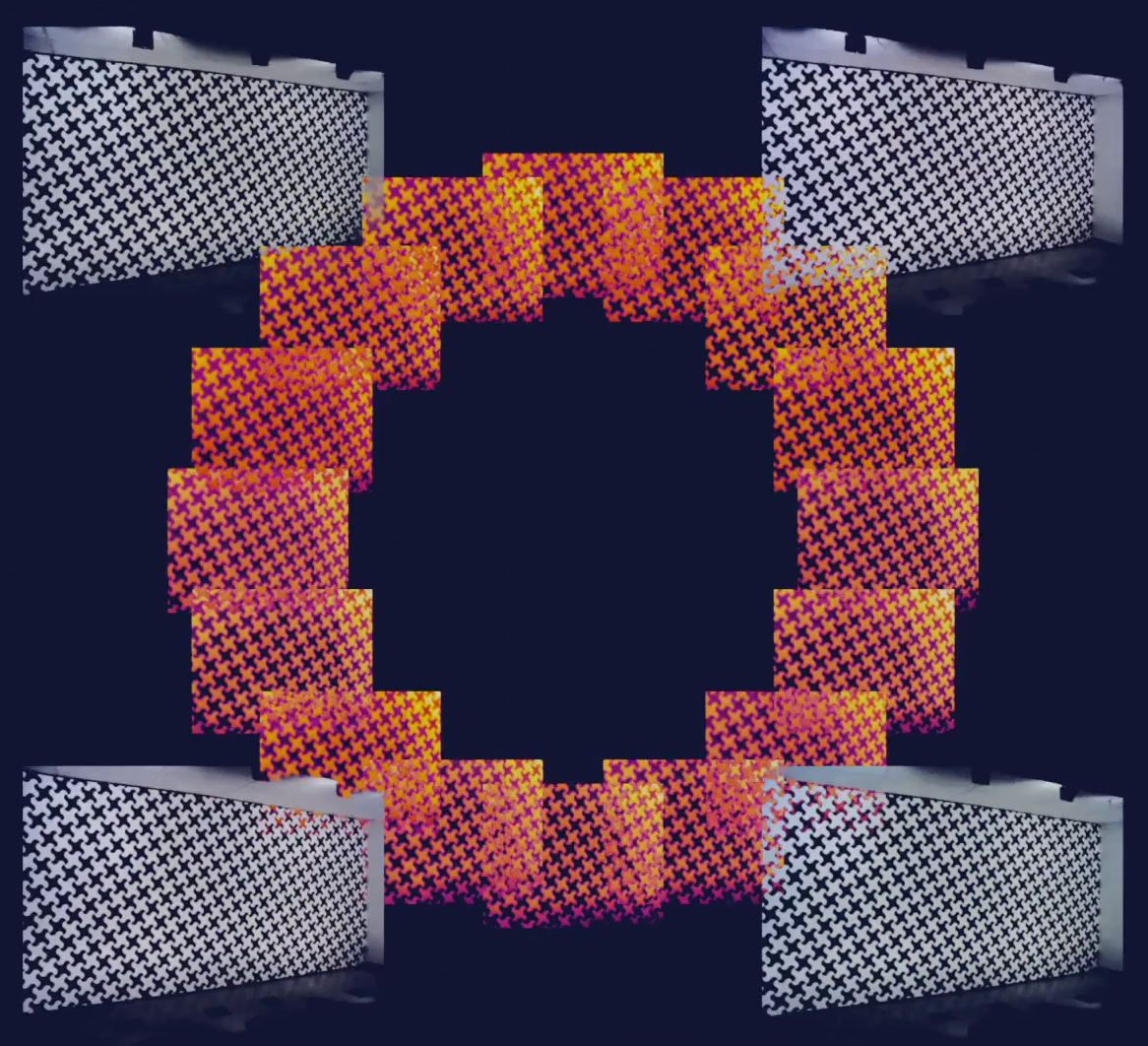}
\captionof{figure}{Images of the target as viewed by 4 RGB and 16 \gls{lwir}
sensors.}
\label{fig:20views}
\vspace{-10pt}
\end{wrapfigure}
\fi

We captured image sets from 4 positions (``stations'') illustrated in
Fig.\ref{fig:stations_plan}, accumulating 700 of them. Off-center stations
(L and R) are needed for accurate measurement of the depth coordinate (target
surface flatness), C is optimal for visible-range cameras \gls{psf} measurements
(wide-angle view of the pattern but far enough for the \gls{dof}), and F is needed for
the \gls{lwir} \gls{psf} measurements that have \gls{dof} limiting minimal
fixed-focus distance to \SI{20}{\meter}. Such a long distance is determined by a
combination of large pixel-size (\SI{12}{\micro\meter}), high lens numerical
aperture (0.7), and a long focal length (\SI{14}{\milli\meter}).

At each station, the goniometer scans the range of $\pm$\SI{40}{\degree}
horizontally and $\pm$\SI{25}{\degree} vertically, so each pattern region covers
each sensor's entire \gls{fov}.

We used all four stations data for
bundle adjustment, then applied station~C data to calculate \gls{psf} for the RGB
cameras (Figure \ref{fig:rgb-psf}) and station~F data to get \gls{lwir}
\gls{psf} (Figure \ref{fig:lwir-psf}). Figure \ref{fig:20views} illustrates the
view of the target by all 20 sensors from station~L, blog
post~\cite{afilippov2021calibration} provides a video illustration of the
calibration procedure.

\iftrue
\begin{wrapfigure}{L}{0.47\textwidth} 
\vspace{-10pt}
\centering \includegraphics[width=0.97\linewidth]{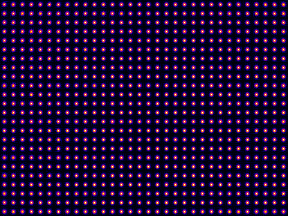}
\captionof{figure}{LWIR Point Spread Function.}
\label{fig:lwir-psf}
\vspace{-15pt}
\end{wrapfigure}
\fi

The first step of processing calibration images is the extraction of the
pattern grid from the individual images. First, software samples images in
reverse binary order, then after finding a suitable candidate, it tries to expand the
found area, adding nodes around. Curved pattern cells help eliminate
false positives when the camera sees pattern reflection on the floor, especially
strong for \gls{lwir} images. After determining the pattern area in
the images, the software ``refines'' found pattern coordinates by generating a
simulated pattern view for each captured image area by warping the ideal
pattern to the second-degree local pattern distortion and calculating 2D
correlation between the captured and simulated areas.

This processing results in a set of subpixel-accurate floating point coordinates
of the pattern grid nodes; we save this data as multi-layer TIFF files (``grid
files'') that are convenient to evaluate manually. 

\iftrue
\begin{wrapfigure}{R}{0.47\textwidth} 
\vspace{-10pt}
\centering \includegraphics[width=0.97\linewidth]{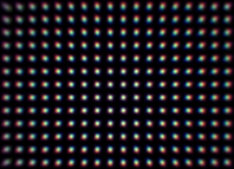}
\captionof{figure}{RGB Point Spread Function.}
\label{fig:rgb-psf}
\vspace{-10pt}
\end{wrapfigure}
\fi

We process grid files generated from the captured images (700 scenes of 20
images each) with an \gls{lma} simultaneously adjusting 3-5 thousand parameters
using 5-10 million points (measured X/Y positions of the pattern grid nodes),
resulting in \gls{rmse} under 0.05~pix. With radial distortion alone (without
non-radial deviations correction), the \gls{rmse} is 0.12-0.15~pix, proving the
more advanced processing required to achieve needed calibration
accuracy, the processing we implemented in our work.

The next step after distortions calculation is measuring the space-variant
\gls{psf} by combining partial arrays, each covering a fraction of the entire
sensor \gls{fov}.
Each image captured from station C for RGB and station F for \gls{lwir}
that contains a sufficiently large view of the target grid is used to calculate
fragments of the whole \gls{psf}, as each image (especially for \gls{lwir} from
station F) covers only a fraction of the entire \gls{fov}. These images are then
combined (first outliers are removed, then remaining are combined in the frequency
domain by separately averaging power spectrum and phase shifts), resulting in
arrays of the space-variant \gls{psf} kernels shown in
Figures~\ref{fig:lwir-psf},\ref{fig:rgb-psf}.

The \gls{psf} kernels have twice higher
resolution than the actual sensors (0.5 pix in each direction), and the distance
between the kernels in these illustrations are not in scale with the image size,
so the \gls{psf} full width in the center is less than \SI{1.6}{\pixel}
compared to \SI{2592x1926}{\pixel} of the full RGB images and
\SI{640x512}{\pixel} of the \gls{lwir} ones. Color channels of RGB images are
processed separately; Figure~\ref{fig:rgb-psf} combines them, revealing
the chromatic aberration.

We then invert the \gls{psf} kernels to get
deconvolution kernels needed for aberration correction of the field-acquired
images.

\subsection{Field Image Acquisition}
\label{subsec:field_activities}

We have captured both daytime dual-modal (\gls{lwir} + \gls{eo}) image sets
(Figure~\ref{fig:footage_lwir-rgb}) and nighttime \gls{lwir}-only ones
(Figure~\ref{fig:footage_lwir}) using our backpack-mounted platform in various
environments (as shown in Figure~\ref{fig:teasedale} and
Figure~\ref{fig:capeblanco}). RGB images from \gls{eo} cameras have a much
higher resolution (\SI{2592x1936}{\pixel} vs.
\SI{640x512}{\pixel}) than \gls{lwir} images, allowing us to use RGB images
as ground truth data for evaluation of the \gls{lwir}-derived depth map accuracy
and training and testing of the \gls{dnn}.

\iftrue
\begin{wrapfigure}{L}{0.4\textwidth} 
\vspace{-10pt}
\centering \includegraphics[width=0.97\linewidth]{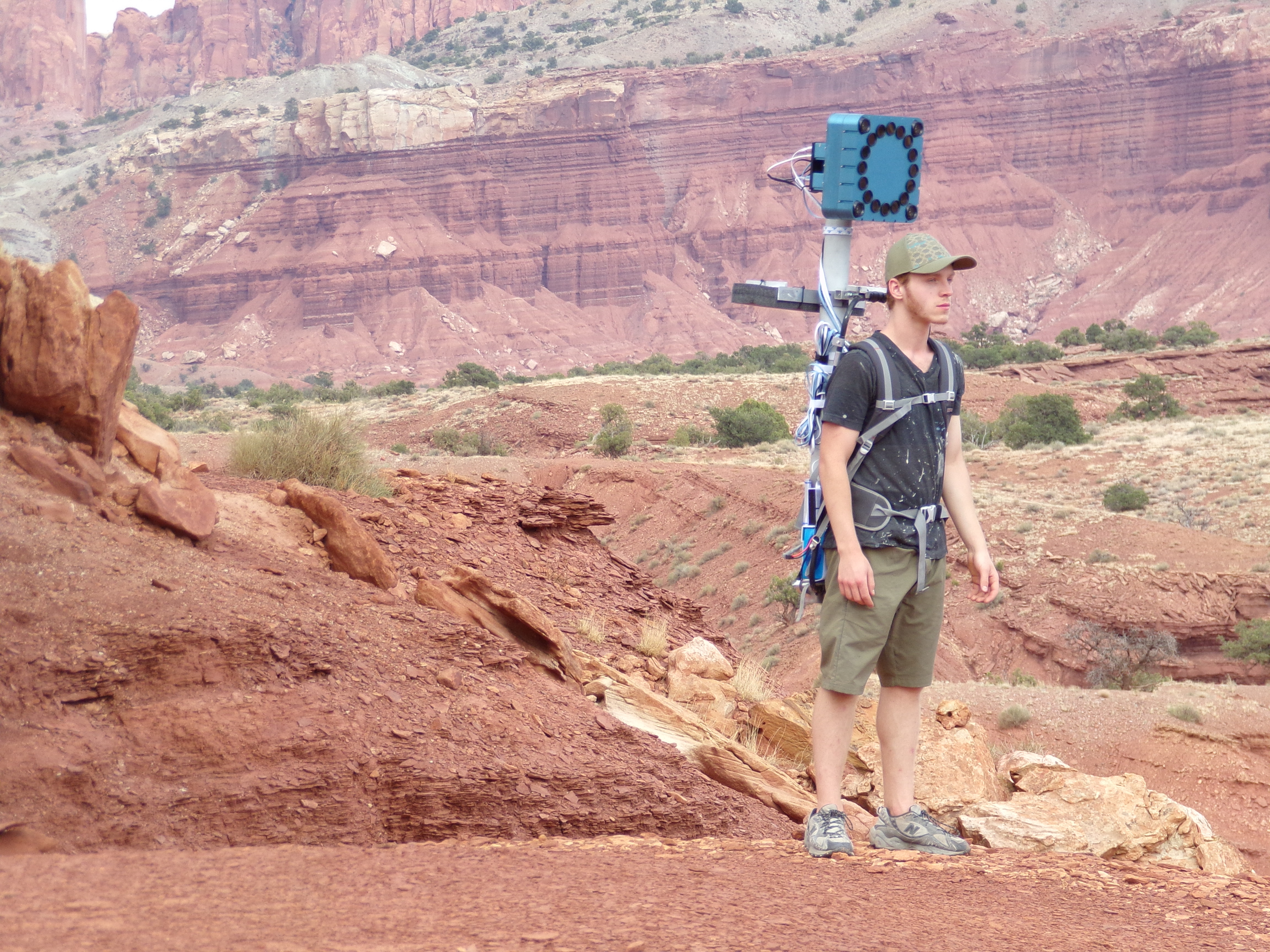}
\captionof{figure}{Image acquisition - the desert.}
\label{fig:teasedale}
\vspace{-10pt}
\end{wrapfigure}
\fi

Due to the limitation of the current camera \gls{fpga} code that lacks
lossless compression of 16-bit thermal images and bandwidth limitations of the
simultaneous 4-channel \gls{lwir} data recording and our need to capture long
image sequences at a full frame rate for interscene contrast improvement, we
implemented a special ``burst'' recording mode. Periodically, each camera module
simultaneously acquires consecutive images at maximal \gls{fps}, filling
corresponding RAM buffers (64~MB per channel). One burst lasts
\SI{1.67}{\second} allowing recording of 100 consecutive \gls{lwir} images  at
\SI{60}{\hertz} (17 images of RGB images recorded at \SI{10}{\hertz}); and
after a \SI{3}{\second} pause, the sequence repeats. We record
the synchronized sequence of JP4-encoded RGB images and 16-bit TIFF \gls{lwir}
images enclosed in 1~GB *.mov files, reducing the number of individual files.

Each recording session (0.5-1.0 hours long) resulted in several hundred
gigabytes of raw data and about a million individual images.
We created preview videos from the raw data files to simplify
selecting initial scene sequences for further processing.
 These files are much smaller snd we were able to
upload them to our web server - they are now available in subfolders of
\href{https://community.elphel.com/files/lwir16/}{https://community.elphel.com/files/lwir16/}.

\iftrue
\begin{wrapfigure}{R}{0.4\textwidth} 
\vspace{-10pt}
\centering \includegraphics[width=0.97\linewidth]{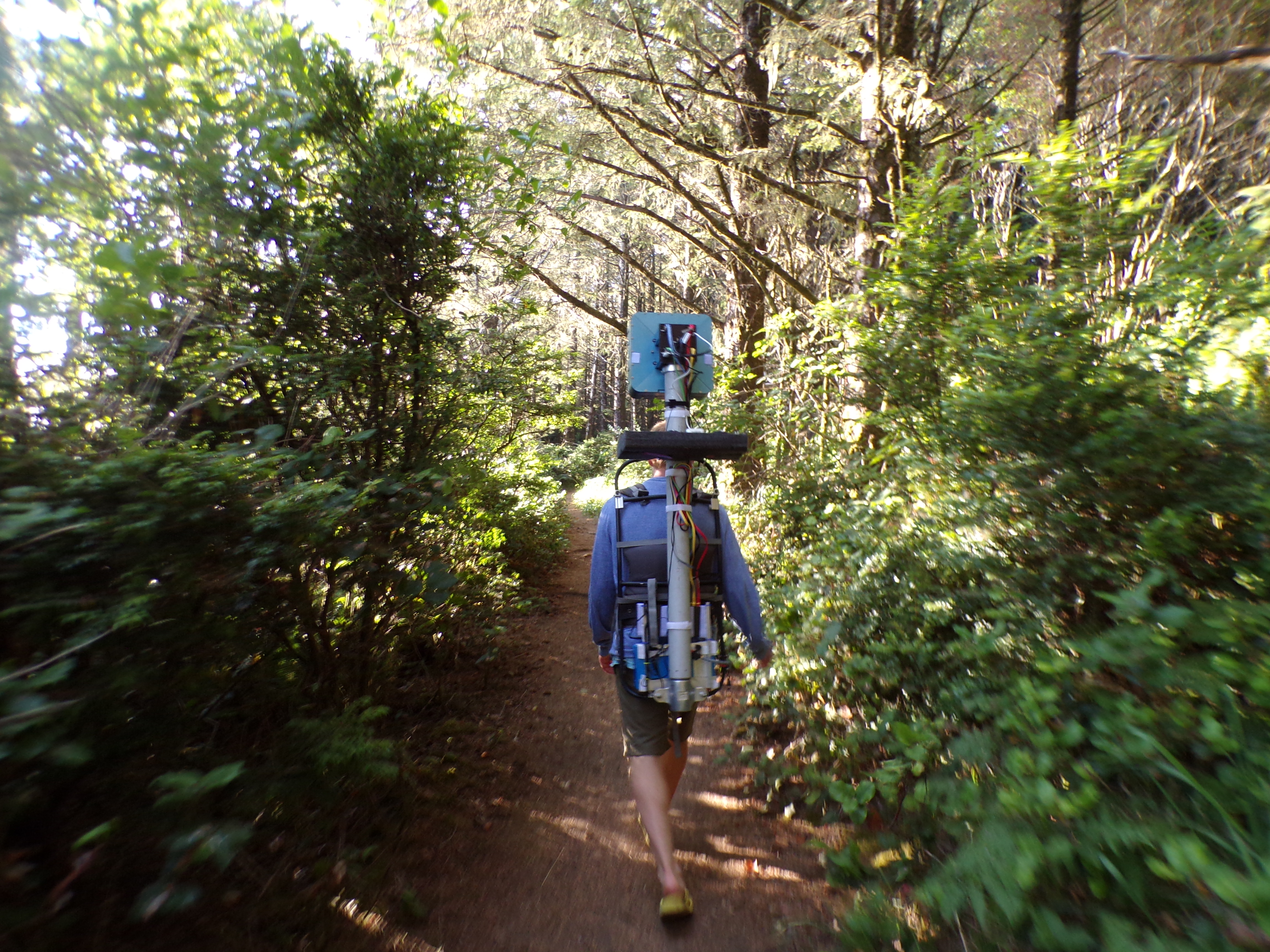}
\captionof{figure}{Image acquisition - the forest.}
\label{fig:capeblanco}
\vspace{-10pt}
\end{wrapfigure}
\fi

Each of the three series contains individual channel videos: chn\_00.webm to
chn\_15.webm for \gls{lwir} channels and chn\_16.webm to chn\_19.webm   for the RGB
ones (RGB channels are available for floras\_lake only, orford and butterfield
are captured at nighttime without any meaningful RGB data). Videos are rendered
at constant 30~fps, so continuously captured segments look as slow motion, and
the missing \SI{3}{\second} gaps result in ``jumps'' in the video.

Each image frame has a visible timestamp that matches the filename of the raw
image. The \gls{lwir} channels numbers start from the top (channel 0);
other channels are located clockwise, looking in the direction parallel to the
camera view. RGB channels are top-left~(16), top-right~(17) bottom-left~(18) and
bottom-right~(19).
In addition to the videos, we created simple viewers to watch stereo-pairs of
images (just captured images without any correction) using Google cardboard
stereo-viewer or a similar device.

\iftrue
\begin{wrapfigure}{L}{0.55\textwidth} 
\vspace{-10pt}
\centering \includegraphics[width=0.97\linewidth]{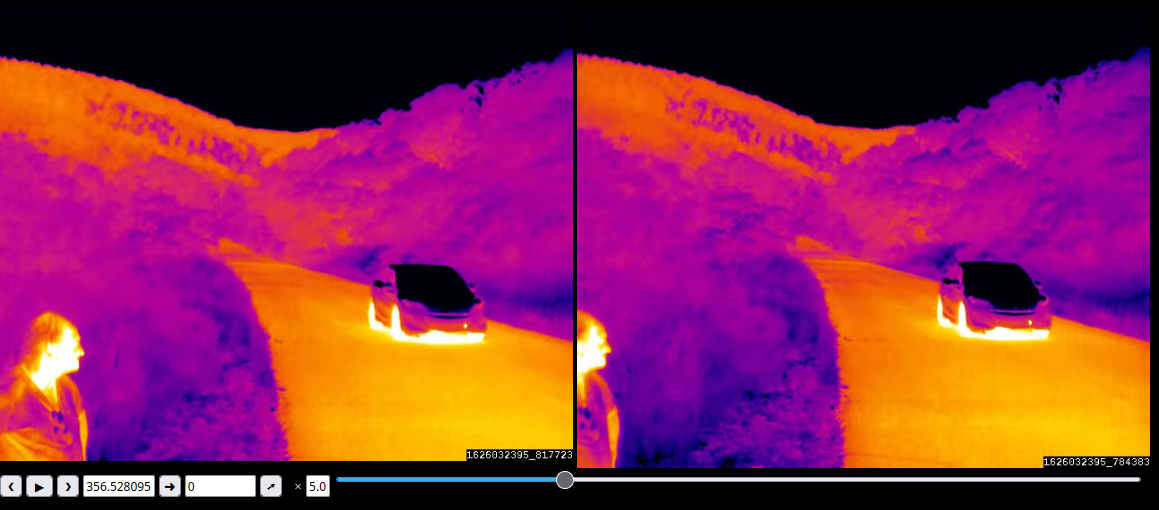}
\captionof{figure}{Preview of the nighttime captured \gls{lwir} footage as a binocular stereo (Butterfield Canyon, UT).}
\label{fig:footage_lwir}
\vspace{-10pt}
\end{wrapfigure}
\fi

Port Orford, Oregon
\begin {itemize}
  \item \href{https://community.elphel.com/files/lwir16/orford/video-1-15.html}{orford/video-1-15.html} --
   two near-top images at approximately the human eye distance (the smallest
   stereo disparity)
  \item \href{https://community.elphel.com/files/lwir16/orford/video-2-14.html}{orford/video-2-14.html}
  \item \href{https://community.elphel.com/files/lwir16/orford/video-3-13.html}{orford/video-3-13.html}
  \item
  \href{https://community.elphel.com/files/lwir16/orford/video-4-12.html}{orford/video-4-12.html}
  -- pair of the most distant cameras (the largest stereo disparity)
\end{itemize}

Butterfield Canyon, Utah
\begin {itemize}
  \item \href{https://community.elphel.com/files/lwir16/butterfield/video-1-15.html}{butterfield/video-1-15.html}
  \item \href{https://community.elphel.com/files/lwir16/butterfield/video-2-14.html}{butterfield/video-2-14.html}
  \item \href{https://community.elphel.com/files/lwir16/butterfield/video-3-13.html}{butterfield/video-3-13.html}
  \item \href{https://community.elphel.com/files/lwir16/butterfield/video-4-12.html}{butterfield/video-4-12.html}
\end{itemize}

\iftrue
\begin{wrapfigure}{R}{0.55\textwidth} 
\vspace{-10pt}
\centering \includegraphics[width=0.97\linewidth]{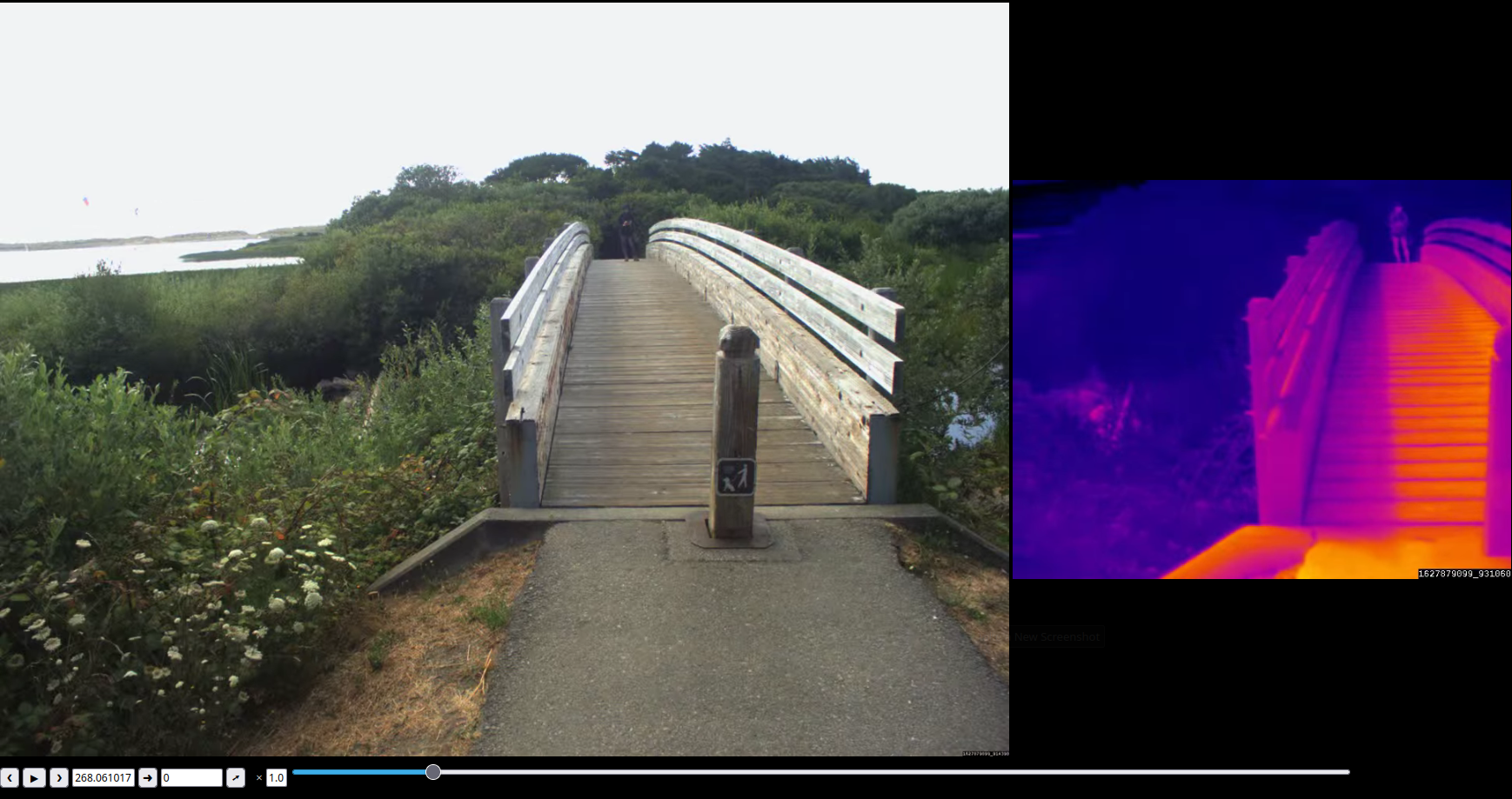}
\captionof{figure}{Preview of the daytime captured footage with RGB and LWIR sensors (Floras Lake, OR).}
\label{fig:footage_lwir-rgb}
\vspace{-10pt}
\end{wrapfigure}
\fi

Floras Lake, Oregon
\begin {itemize}
  \item \href{https://community.elphel.com/files/lwir16/floras_lake/video-1-15.html}{floras\_lake/video-1-15.html}
  \item \href{https://community.elphel.com/files/lwir16/floras_lake/video-2-14.html}{floras\_lake/video-2-14.html}
  \item \href{https://community.elphel.com/files/lwir16/floras_lake/video-3-13.html}{floras\_lake/video-3-13.html}
  \item \href{https://community.elphel.com/files/lwir16/floras_lake/video-4-12.html}{floras\_lake/video-4-12.html}
  \item \href{https://community.elphel.com/files/lwir16/floras_lake/video-17-16.html}{floras\_lake/video-17-16.html}
   -- a stereo-pair of RGB videos (frame size is twice reduced, each frame is
   repeated six times to match \gls{lwir})
  \item \href{https://community.elphel.com/files/lwir16/floras_lake/video-14-16.html}{floras\_lake/video-14-16.html}
   -- not a stereo-pair, just a side-by-side comparison of RGB and \gls{lwir}
   videos.
\end{itemize}

We created these images for scenes preview and selection only; they are not
a part of the image processing flow.

\needspace{15\baselineskip}

\subsection{Extending Software to Support Arbitrary Number of Sensors}
\label{sub:soft-arbitrary-sensors}

Our previously developed image-processing and 3D-reconstruction software that we
have been developing since 2017 specifically depended on a quadocular camera
configuration with four sensors located at the corners of a square. Such layout
allowed the simple fusion of the horizontal stereo-pairs with the
vertical ones by a lossless transposing of the 2D phase correlation outputs. For
the larger number of cameras, we needed rotation by almost arbitrary angles and
scaling of the phase correlation outputs to combine results of the
correlations with various baselines -- from  1/5 of the diameter ($sin(\pi/16)$)
for the neighboring sensors to the full diameter. We accomplished this by
developing different methods of multi-pair consolidation of the 2d correlation
and refactoring the code to support an arbitrary number of sensors.

One of the first functions of the updated code tested for 16-sensors scenes
(simultaneously captured sets of \gls{lwir} images) was the code for field
calibration (``Lazy Eye'' correction -
\href{https://git.elphel.com/Elphel/imagej-elphel/blob/lwir16/src/main/java/com/elphel/imagej/tileprocessor/ExtrinsicAdjustment.java}
{ExtrinsicAdjustment.java}) to measure and compensate for minor misalignment of
the sensor modules that may develop over time or caused by vibrations or
mechanical distortions caused by temperature variations. While we did not yet
process a massive number of captured scenes to calculate statistical properties
of the required correction amounts, selective processing of the image sets captured
at different locations (described in subsection~\ref{subsec:field_activities})
confirmed that the misalignment of the sensors is below \SI{0.1}{\pixel} and
verified the overall mechanical stability of the prototype system
(Figure~\ref{fig:prototype}).

\subsection{Generating a Single-Scene Disparity Map}
\label{sub:single-scene-map}

Single-scene generation of the disparity map with the multi-sensor camera is an
iterative process that resembles human stereo-vision involving eye convergence
followed by processing the residual mismatch between the two registered images.
In our case, pre-shifting corresponding image tiles before calculating the 2d
phase correlations replaces mechanical eyes movement.
the mechanical movement of the eyes is substituted by pre-shifting corresponding
image tiles before calculating the 2d phase correlations.
The calculation of the pairwise correlation between corresponding image tiles
requires significant overlap between the registered images, and the correlation
result quickly degrades when tile overlap reaches its half-width.
The iterative process starts from some initial approximation of the disparity
map referenced to a virtual camera positioned in the center of the sensor
circle.
According to the \gls{difrec} method, this virtual camera has the optical
distortions averaged from the measured during factory calibration distortions of
the sensors.
The initial approximation may be obtained by a full disparity sweep, prediction
from the previous scene, expanding from the known areas, or other algorithms.
These initial generalized disparity values are reciprocal to the object
distances for each tile center on a uniform grid (stride 8) of the virtual
camera.
For the traditional binocular stereo, it will be the actual shift between the
two images in pixels. In the multi-sensor case, this single for each tile scalar
value is used to calculate a vector that indicates how each image should be shifted to
match the other ones. For a circular camera configuration, this vector
has the same length with the direction matching the camera's location.
The exact vector calculation uses the calibration results and compensates for
measured misalignments and distortion differences.

\iftrue
\begin{wrapfigure}{L}{0.57\textwidth} 

\vspace{-10pt}
\centering \includegraphics[width=0.97\linewidth]{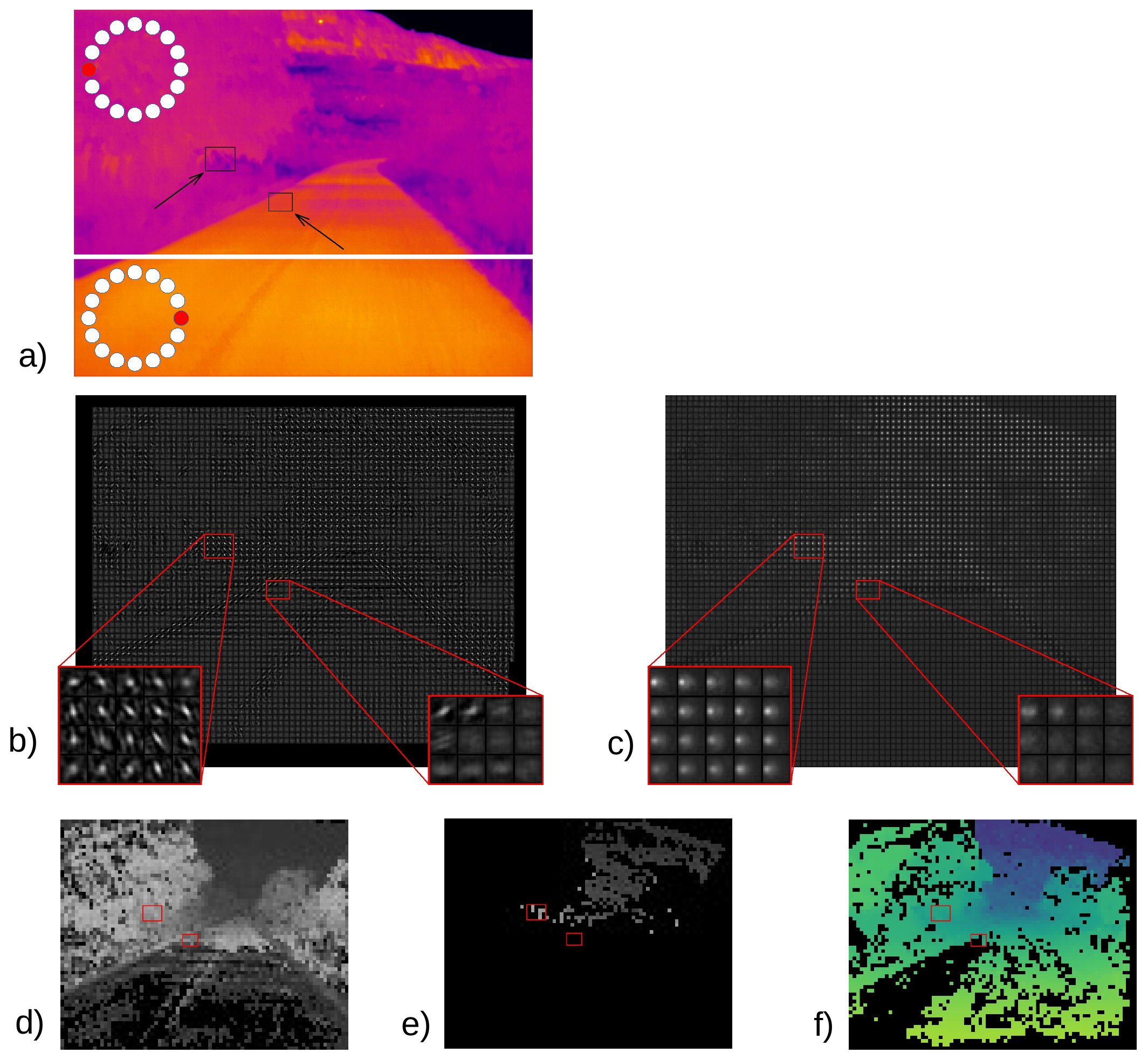}
\captionof{figure}{Single-scene disparity map generation:
a) source thermal images; b) phase correlations for an image pair converged at
infinity; c) multi-pair combination of phase correlations; d) disparity map by
center-of-mass argmax of c); e) disparity map by \gls{lma} argmax of b); f)
full disparity map by \gls{lma} for disparity sweep.}
\label{fig:intrascene-map}
\vspace{-10pt}
\end{wrapfigure}
\fi
Knowing these 16 vectors, we implement the required pre-shift in two steps:
\begin {itemize}
  \item round the shift to the nearest integer and use it to select a
  $16\times16$ corresponding tile in each source image, and
  \item apply the residual \SI{\pm0.5}{\pixel} shift as a lossless rotation in
  the \gls{fd}.
\end{itemize}

The \gls{tp} operates on each tile independently (and in parallel when
implemented in \gls{gpu} or custom hardware), starting with the conversion to
the \gls{fd}, correction of the optical aberrations by the deconvolution with the
known from the factory calibration kernels, applying calculated fraction-pixel
shift by complex values rotation and calculating the 2d phase correlations.
In the current work, we calculated all 120 sensor pairs, leaving the reduction
of this number for future effort. The calculation of the phase correlation involves
the normalization of the \gls{fd} representation by dividing each complex value
of the 2d array by its absolute value. In the noiseless case, this would lead to
a sharp 2d $\delta$-function when converted back to the pixel domain, but it
would amplify noise and make it difficult to find the correlation maximum and
thus the disparity. This noise amplification is mitigated by adding a small
constant to the absolute value (a ``fat zero'') in the denominator during
normalization, balancing between low noise/wide maximum (regular correlation for
a high value of the fat zero) and high noise/sharp maximum (pure phase
correlation for ``fat zero'' equals zero).

Calculation of the correlation maximum ($argmax$ function) determines the
relative distance to the object (a voxel of the depth map); this residual
disparity is added to the initial disparity estimation. The process 
repeats until the change in disparity falls below a specified threshold.
This iterative procedure provides high accuracy of the disparity calculation
achieving deep subpixel resolution because while pre-shifting in the \gls{fd}
is lossless, measurement of the residual disparity is prone to  pixel-locking (Fincham and
Spedding~\cite{fincham1997low}) effect that distorts the results. 

Figure~\ref{fig:intrascene-map} illustrates a single-scene disparity map
calculation. The 16 thermal images of \ref{fig:intrascene-map}a provide input;
the horizontal split shows a disparity between the images from the leftmost
sensor (top) and the rightmost (bottom).
The two black rectangles indicate the high-contrast and low-contrast areas.
Figure~\ref{fig:intrascene-map}b shows one of the 120 pairwise phase
correlations.
When the tile corresponds to an object at the same
distance, each of its 120 correlations has the
elliptic shape of the same size/orientation with its center shifted from
the tile center proportional to the vector that connects lens centers. One of the simple
methods that combine multiple correlation tiles results in
\ref{fig:intrascene-map}c. Each partial correlation output is rotated and scaled
to align the disparity vectors, and then accumulated. The result loses shape
information (helpful in following object edges) but provides an easy
method to combine correlation outputs and increase \gls{snr} before looking for
an argmax.

As the \ref{fig:intrascene-map}b is calculated with zero pre-shift (``eyes'' are
converged at infinity) the \ref{fig:intrascene-map}c outputs meaningful results
only for distant objects with the disparity of less than approximately 4 pixels.
The \ref{fig:intrascene-map}d shows this residual disparity measured by a
center-of-mass method applied to \ref{fig:intrascene-map}c and results in false
disparity for the highway pavement in the bottom part of the image (as if it is
much farther than the pavement representation above it). The
\ref{fig:intrascene-map}e outputs the result of the \gls{lma} fitting applied
directly to all 120 partial correlations of \ref{fig:intrascene-map}c (bypassing
averaging in \ref{fig:intrascene-map}c). It uses more pessimistic filtering that
discards low confidence tiles and requires fewer iterations to achieve the same
depth accuracy.
The last subplot -- \ref{fig:intrascene-map}f provides a full disparity map
resulting from the iterative process of disparity refinement after the initial
disparity sweep.
It also tries to fill the gaps caused by the low-contrast areas by reducing
lateral resolution and consolidating multiple neighboring tiles, assuming that
low-contrast usually assumes that the corresponding surface in 3d is flat or
near-flat. There are still some gaps in the disparity map (black) remaining,
and the following subsection describes how to boost contrast by interscene
accumulation of the intrascene correlation results.

\subsection{Generating a Disparity Map from a Scene Sequence}
\label{sub:multi-scene-map}

\iftrue
\begin{wrapfigure}{R}{0.6\textwidth} 
\vspace{-10pt}
\centering \includegraphics[width=0.97\linewidth]{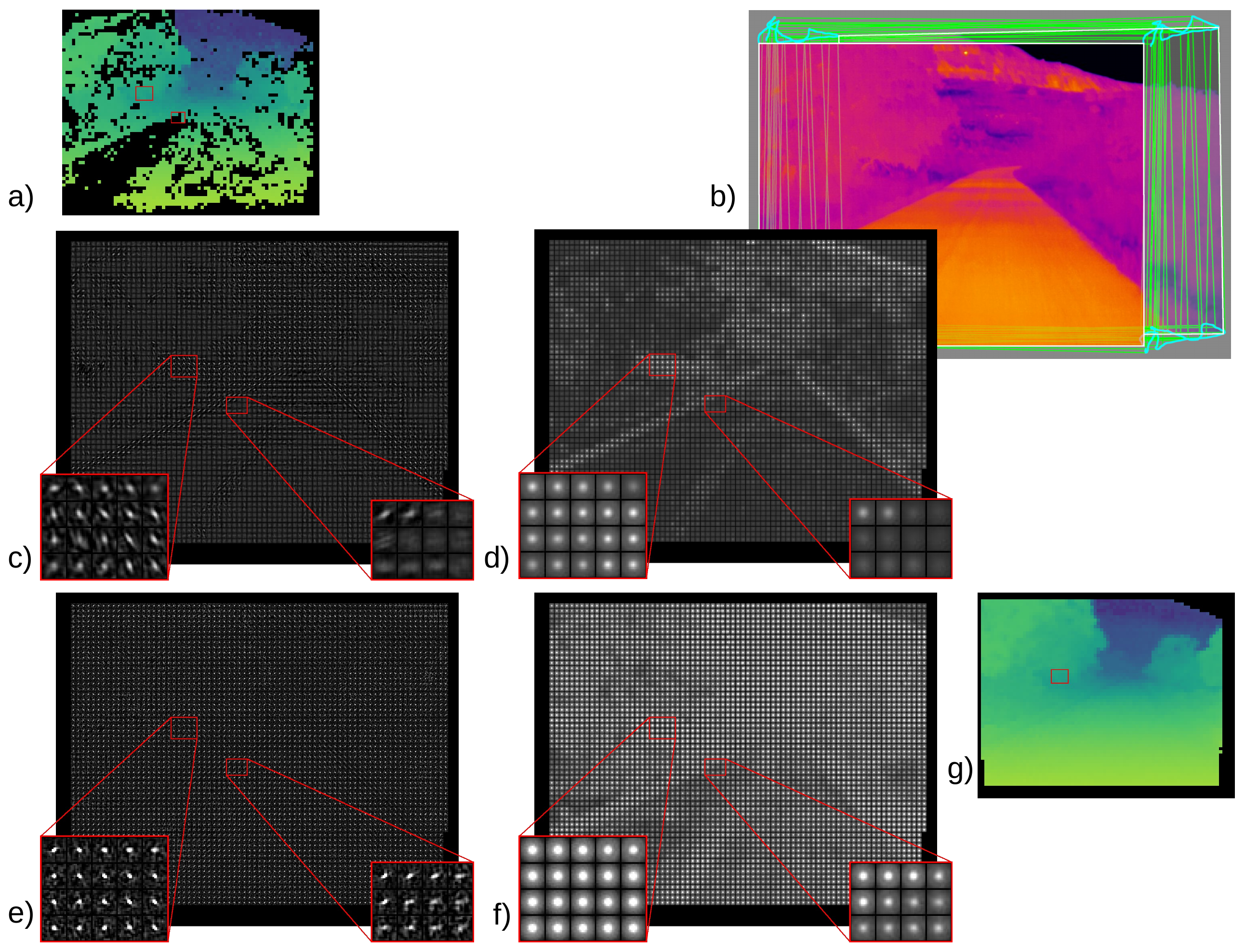}
\captionof{figure}{Multi-scene disparity map generation:
a) disparity map from the last (reference) scene in a sequence; b) outlines
of the frames in a sequence; c) single-scene individual pair 2d correlations;
d) single-scene multi-pair combination of phase correlations; e) single-pair
combination of 2d correlations from all the 99 scenes; f) multi-scene multi-pair
combination of phase correlations; g) disparity map generated from a scene
sequence of the moving camera.}
\label{fig:interscene-map}
\vspace{-10pt}
\end{wrapfigure}
\fi

The interscene accumulation is based on an observation that most of the
low-contrast objects of the scene, such as rocks, bushes, potholes on the road,
are static in 3D, while the moving objects (people, animals, machinery) have
high thermal contrast that allows their robust detection and ranging. There are
certain situations where moving objects may also have low thermal contrast, and
we plan to implement interscene accumulation of such objects in Phase~II by
extending image processing capabilities to compensate for their movement.

The final real-time implementation will optimize processing to reduce the number
of calculations for each newly acquired scene (use incremental 3D model
generation). However, the current research only aims to enhance contrast and
depth accuracy of the last acquired (reference) scene number 99, using previous scenes (1 to 98) for
that purpose.

The interscene accumulation process starts with building partial disparity maps
for all scenes in a sequence described in the previous
Subsection~\ref{sub:single-scene-map}.
Figure~\ref{fig:interscene-map}a illustrates the partial disparity map of a
reference scene.
The software scans all the subsequent scene pairs: reference~(99) and the
previous 98, 98 and 97, up to 2 and 1.
It fits each pair's partial disparity map using per-tile 2D correlations
(reusing the same low-level methods as for image correlation) and results in the
camera linear movement and rotation 6-element vector.
At this iteration, we only compare subsequent scenes (1/60 of a
second apart) as they differ little from each other, making correlation simple.
For matching, we use each scene ``disparity confidence'' available for
each scene in addition to the disparity itself and an output from the high-pass
filter from the down-scaled source thermal images. During this iterative
process, we re-calculate the predicted appearance of the other scene from the
reference scene camera position, use per-tile correlation to measure X and Y
offsets for each tile, and then fit a 6-element global camera movement vector
with \gls{lma}.

Having built pairwise scene camera egomotion vectors, we perform a second pass
to calculate relative camera positions at each scene to the reference one (97 to
99, 96 to 99, \dots, 1 to 99). Just a multiplication of the pairwise matrices
would accumulate errors, but we can use them in the iterative process.
Consider, we have already referenced scene 51 to the 99. 
We can multiply the available pairwise movement matrix from scene 50 to 51 and
from 51 to 99, then use it as an initial approximation of 50 to 99 movement and
improve accuracy by the direct correlation between scenes 50 and 99, and repeat this algorithm until all
scenes camera positions are referenced to the scene~99.
Figure~\ref{fig:interscene-map}b illustrates this by showing a stack of
offset scene images under reference scene one.

Individual depth maps \ref{fig:interscene-map}a are filtered and averaged
(avoiding averaging the foreground with background elements) with disparity gaps
filled by predictions resulting in a dense (defined for every tile of the
reference scene) but maybe inaccurate disparity map to be enhanced later by the
simultaneous correlations processing.

Then we calculate pairwise correlations for each scene, but this time each image
tile center's locations are calculated to match reference scene tiles, not the
tiles of the virtual image of that scene as we did before.
Figures~\ref{fig:interscene-map}c and d represent single-pair phase correlations
and their combinations, respectively, similar to earlier
Figures~\ref{fig:intrascene-map}b and c. The differece is that this time the
full-scene pre-shift based on the known disparity map is applied, so each tile
phase correlation output is centered.
The later steps do not use single-scene \ref{fig:interscene-map}b and c; they
are only provided for comparison.
Instead, the \gls{fd} representation of the correlation (still as complex
values, not collapsed by the transformation to the pixel domain) are averaged
among the same tile pair in all scenes where the corresponding object is visible
(it is present in the \gls{fov} and not occluded).
We apply the non-linear step of amplitude normalization followed by the inverse
transformation only once after averaging preserving \gls{snr} increase as a
square root of the number of independent measurements even when it is much less
than 1.0 in the individual scenes.

\iftrue
\begin{wrapfigure}{L}{0.45\textwidth}
\centering \includegraphics[width=0.97\linewidth]{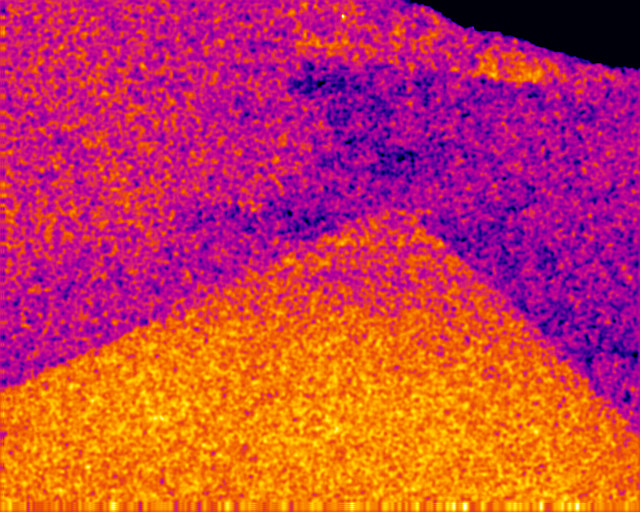}
\captionof{figure}{Captured image + noise.}
\label{fig:lwir_with_noise}
\vspace{-15pt}
\end{wrapfigure}
\fi

Figure~\ref{fig:interscene-map}e illustrates individual pair phase
correlations and ~\ref{fig:interscene-map}f) -- a combination of all 120 pairs
for a center-of-mass argmax calculation. More accurate \gls{lma} fitting uses 
individual pairs of ~\ref{fig:interscene-map}e. We will feed
these shape-preserving per-pair phase correlation outputs to the
\gls{dnn} as we did with a quad-camera in previous research.

\subsection{Evaluating Contrast Gain}
\label{sub:evaluating-contrast-gain}

When we obtained dense disparity maps using captured thermal images from all
available (16) sensors, we needed to evaluate the influence of the number of
sensors and the contribution of interscene accumulation.
Setting up a ``fair competition" between different sensor configurations is not a
trivial task because the complete procedure of calculating the depth map from
the sequence of the captured thermal images involves multiple non-linear
processing stages depending on a large number of parameters. It would be
extremely difficult to explore and optimize all the parameter space for each of
them and to guarantee that we are comparing optimized for binocular to optimized
for quadocular, and the same for a higher number of sensors.
To resolve this challenge, we tried to narrow the ``competition'' task and
reduce the number of independent parameters that influence the quality of the
depth map output.

\iftrue
\begin{wrapfigure}{R}{0.67\textwidth} 
\vspace{-10pt}
\centering \includegraphics[width=0.97\linewidth]{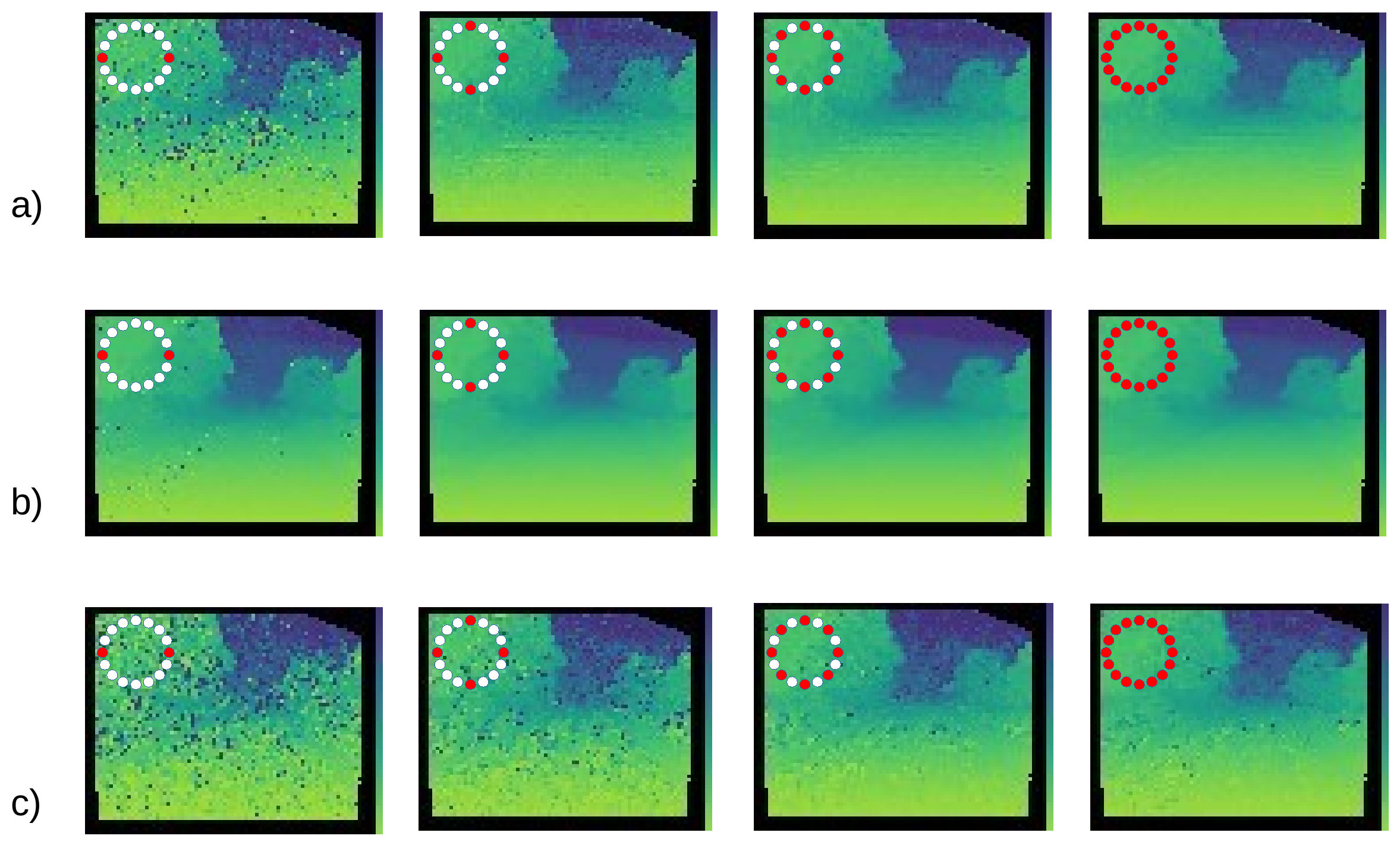}
\captionof{figure}{Depth map with interscene accumulation: a) single-scene
disparity maps for 2, 4, 8, and 16 sensors; b) 99-scene disparity maps; c)
99-scene disparity maps generated from the source images with added uncorrelated
synthetic noise.}
\label{fig:processed-2-4-8-16}
\vspace{-10pt}
\end{wrapfigure}
\fi

We started with the best depth map we could achieve using all scenes and
all sensors, considering it to be the best approximation of the ground truth.
Then we added the same constant value (it is currently 1.4142 pixels) to all
tiles and used this offset depth map as initial pre-shift (equivalent to ``eye
convergence'') before calculating the first iteration of the per-tile phase
correlation. Then we performed a fixed number (10) of disparity refinement
iterations of measuring residual disparity from argmax of combined correlations,
adding it to disparity estimation (pre-shift) and re-calculating correlations.

Each tile disparity value was either converging to a final value (last step
typically under \SI{1e-5}{\pixel}) or diverging.
Furthermore, if converging, the disparity could be close to the expected value
(obtained from 99 scenes, 16 sensors) or far.
We considered tiles with the final error above \SI{2.0}{\pixel} as ``diverged''
(reducing calculated map density) and calculated \gls{rmse} only for the
converged ones.

For this test, we used only center-of-mass argmax as it does not depend on any
configuration parameters, while the more advanced \gls{lma}-based method depends
on them.
The center-of-mass method provides ``confidence'' output, but its interpretation
depends on different heuristic thresholds for different configurations, so we
did not filter depth map outputs from outliers, visible in
Figure~\ref{fig:processed-2-4-8-16}.
The only remaining parameter that influences the result is phase correlation
``fat zero,'' and we verified that changing it twice (both ways) from the
selected value did not significantly affect any configurations.
The amount of initial offset (1.4142) has to be significantly smaller than
minimal tile overlap ($\lessapprox 4$) and larger than the maximal \gls{rmse} we
expected to measure.

\iftrue
\begin{wrapfigure}{L}{0.47\textwidth}
\vspace{-10pt}
\centering \includegraphics[width=0.97\linewidth]{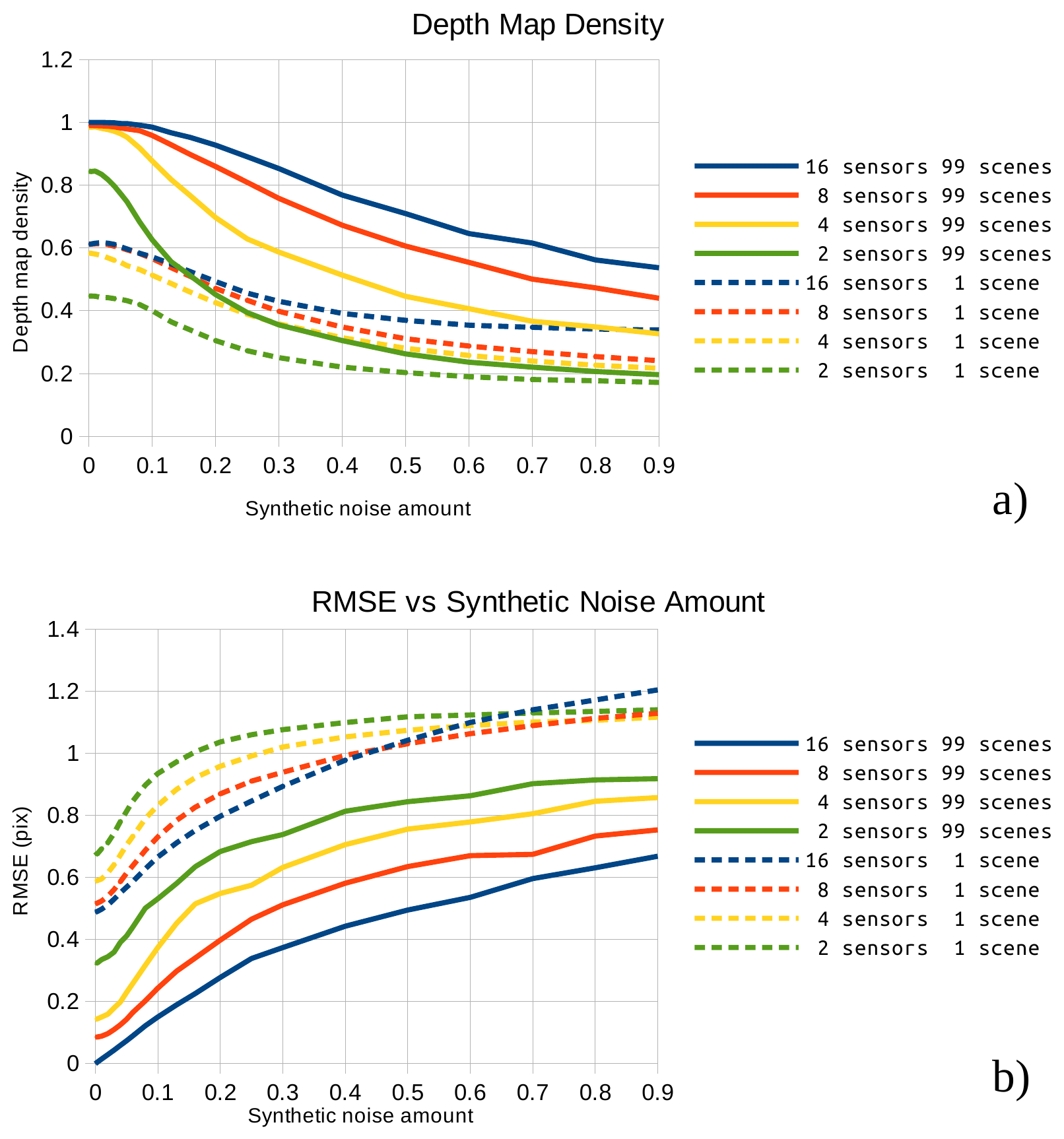}
\captionof{figure}{Disparity map quality vs. added synthetic noise:
a)\gls{rmse}; b) density.}
\label{fig:rmse-density-graphs}
\vspace{-10pt}
\end{wrapfigure}
\fi

Figure~\ref{fig:processed-2-4-8-16}a shows the resulting depth maps for
single-scene 2, 4, 8, and all 16 sensors; \ref{fig:processed-2-4-8-16}b -- same
for interscene accumulation. The thermal images we used were acquired only two
hours after sunset and had significant contrast, so the interscene results
(Figure~\ref{fig:processed-2-4-8-16}b were almost identical for 4, 8, and 16
sensors. To reveal the difference, we added synthetic noise of variable
amplitude to the source images, such as in Figure~\ref{fig:lwir_with_noise}
(amplitude of 0.3 of the entire pixel range captured), and used it to generate 
Figure~\ref{fig:processed-2-4-8-16}c.

The software generated synthetic noise images for each scene, each sensor
once, saved and then mixed to the captured images, so each sensor configuration
got the same input data.

We used the following simplified model to compare the performance of different
sensor configurations quantitatively.
Each sensor has the same intrinsic noise, invariant of the pixel position in the
field of view.
The 3d scene element corresponding to each tile of the reference scene defines
its useful signal.
Different sensor configurations have different signal gains, and the outcome of
the disparity calculation for each tile depends on the \gls{snr} for that tile.
During this procedure, we applied different amounts of synthetic noise,
measuring depth map density and \gls{rmse}. Depth map density shown in
Figure~\ref{fig:rmse-density-graphs}a is a fraction of all tiles for which the
repetition of ten cycles of disparity refinement converged (to less than
\SI{0.01}{\pixel}) and differed from the expected value by less than
\SI{2.0}{\pixel}. The \gls{rmse} shown in Figure~\ref{fig:rmse-density-graphs}a
excludes diverged tiles.
The single-scene graphs (plotted with dashed lines) represent averaging of 16
instances of noise files of the same amplitude and have less random
fluctuations than those for the interscene results (solid lines), each
representing just a single set of noise files (because of very high calculation
time).

According to the used model, the individual plots should differ
by a horizontal scaling factor with the offset (for noise squares) of the
intrinsic noise of the physical sensor that is still present when there is no
synthetic noise.
The simplified model is not very accurate, \eg the \gls{rmse} plot for 16-sensor interscene
intersects that of 8-sensor one at the relative noise level of 0.4 while the
corresponding density falls at a lower rate than for the other number of
sensors.
That can be explained as that higher number of sensors locks to a larger variety
of different objects correlation tiles present in the scene and fail gracefully
by just an increased \gls{rmse} rather than by a complete divergence.

\iftrue
\begin{wrapfigure}{R}{0.68\textwidth}
\vspace{-10pt}
\centering \includegraphics[width=1.0\linewidth]{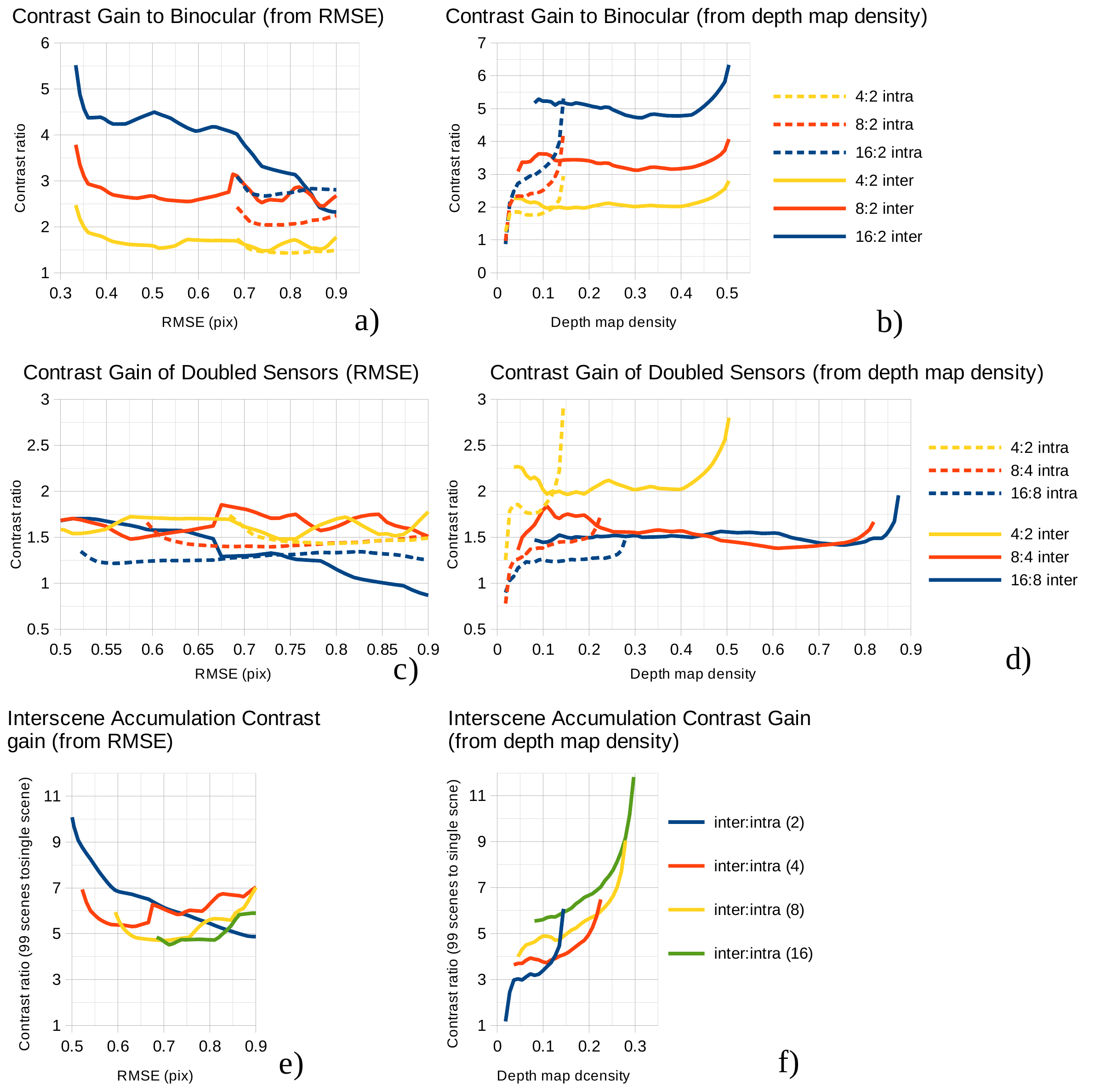}
\captionof{figure}{Contrast gains.}
\label{fig:gain-ratios}
\vspace{-10pt}
\end{wrapfigure}
\fi

Figure~\ref{fig:gain-ratios} represents ratios of the inverted functions of
Figure~\ref{fig:rmse-density-graphs}, ratios of the synthetic noise levels for
the same values of the \gls{rmse} (a, c, and e), and map density (b, d, and
f ). The result contrast improvements for different numbers of sensors is
calculated from combined data of \ref{fig:rmse-density-graphs}c and
\ref{fig:rmse-density-graphs}d and the interscene gain -- from
\ref{fig:rmse-density-graphs}e and \ref{fig:rmse-density-graphs}f by
averaging the contrast ratios (vertical axes).
The amount of intrinsic sensor
noise (optimal value is 0.044) is determined by minimizing the spread of the
contrast gains. 

We expected the intrascene contrast gain to be proportional to a square root of
the independent number of stereo-pairs; this number is 1 for binocular and
equals the number of sensors for other configurations where each sensor image
contributes to two orthogonal stereo-pairs.  These orthogonal pairs are
independent measurements of the average disparity as for each pair matching only
image shift parallel to the corresponding baseline contributes to the output.
The measured results (presented in Table~\ref{tab:contrast-gain}) for intrascene
contrast gain are close to the predicted.
The interscene accumulation should result in up to a square root of the number
of measurements ($10\times$ for 99 scenes), and so far, we achieved only
$5.5\times$ gain.
Part of the lower gain is caused by insufficient scene overlap that reduces
the number of measurements per tile.
Another part may be caused by the
manufacture-proprietary image processing inside the sensor that introduces
a dependency between consecutive images.
We will conduct more experiments to pinpoint this discrepancy's reasons.


\begin{longtable}{|p{0.65\textwidth}|p{0.1\textwidth}|p{0.1\textwidth}}
\caption{Contrast Gain over Single-Scene Binocular Stereo}
\label{tab:contrast-gain} \\
\hline
  \multicolumn{1}{|c|}{
    \textbf{Configuration}
  }&
  \multicolumn{1}{ c|}{
    \textbf{Measured Gain}
  }&
  \multicolumn{1}{ c|}{
    \textbf{Predicted Gain}
  }\\
\hline
  \multicolumn{1}{|l|}{
    4-sensor over binocular
  }&
  \multicolumn{1}{ c|}{
    1.79 
  }&
  \multicolumn{1}{ c|}{
    2.0
  }\\
\hline
  \multicolumn{1}{|l|}{
    8-sensor over binocular
  }&
  \multicolumn{1}{ c|}{
    2.72
  }&
  \multicolumn{1}{ c|}{
    2.83
  }\\
\hline
  \multicolumn{1}{|l|}{
   16-sensor over binocular
  }&
  \multicolumn{1}{ c|}{
    3.84
  }&
  \multicolumn{1}{ c|}{
    4.0
  }\\
\hline
  \multicolumn{1}{|l|}{
   99-scene over single scene
  }&
  \multicolumn{1}{ c|}{
    5.5
  }&
  \multicolumn{1}{ c|}{
    9.95
  }\\
\hline
  \multicolumn{1}{|l|}{
   99-scene, 16 sensor over single scene binocular
  }&
  \multicolumn{1}{ c|}{
    21.1
  }&
  \multicolumn{1}{ c|}{
    39.8
  }\\
\hline
\end{longtable}

Using measured contrast values, we can calculate the
effective system \gls{netd} -- required thermal sensitivity of a pair of
hypothetical thermal sensors in a traditional binocular stereo configuration
(without interscene accumulation) to achieve the same depth map density and
accuracy as a multi-sensor system build of \SI{40}{\milli\kelvin} \gls{lwir}
image sensors. 
 
\begin{longtable}{|p{0.65\textwidth}|p{0.1\textwidth}|p{0.1\textwidth}|p{0.1\textwidth}|p{0.1\textwidth}}
\caption{Effective System \gls{netd}}
\label{tab:system-netd} \\
\hline
  \multicolumn{1}{|c|}{
    \textbf{Number of Scenes}
  }&
  \multicolumn{1}{ c|}{
    \textbf{Binocular}
  }&
  \multicolumn{1}{ c|}{
    \textbf{4-Sensor}
  }&
  \multicolumn{1}{ c|}{
    \textbf{8-Sensor}
  }&
  \multicolumn{1}{ c|}{
    \textbf{16-Sensor}
  }\\
\hline
  \multicolumn{1}{|c|}{
    1
  }&
  \multicolumn{1}{ c|}{
    40.0 mK
  }&
  \multicolumn{1}{ c|}{
    22.4 mK
  }&
  \multicolumn{1}{ c|}{
    14.7 mK
  }&
  \multicolumn{1}{ c|}{
    10.5 mK
  }\\
\hline
  \multicolumn{1}{|c|}{
    99
  }&
  \multicolumn{1}{ c|}{
    7.3 mK
  }&
  \multicolumn{1}{ c|}{
    4.1 mK
  }&
  \multicolumn{1}{ c|}{
    2.71 mK
  }&
  \multicolumn{1}{ c|}{
    1.9 mK
  }\\
\hline
\end{longtable}





\bibliographystyle{ieeetr} 

\needspace{3\baselineskip}
{\small
\bibliography{elphel-bib-glossary/elphel}

\begin{thebibliography}{1}

\bibitem{filippov2019method}
A.~Filippov, ``Method for the {FPGA}-based long range multi-view stereo with
  differential image rectification,'' Apr.~28 2020.
\newblock US Patent 10,638,109 B2.

\bibitem{afilippov2021calibration}
A.~Filippov, ``Calibration of the {LWIR16} camera prototype.''
  \url{https://blog.elphel.com/2021/07/calibration-of-the-lwir16-camera-prototyp},
  2019.
\newblock Technical blog post.

\bibitem{fincham1997low}
A.~Fincham and G.~Spedding, ``Low cost, high resolution {DPIV} for measurement
  of turbulent fluid flow,'' {\em Experiments in Fluids}, vol.~23, no.~6,
  pp.~449--462, 1997.

\end{thebibliography}
}



\section {Acronyms List}
\vspace{-7pt} 
\begin{multicols}{2}
\printglossary[style=index,nogroupskip=true,title={}] 
\end{multicols}

 \ReportDate{11--19--2021}
 \ReportType{Final}
 \DatesCovered{May 2021--November 2021}
 \Title{Long-Range Thermal 3D Perception in Low Contrast Environments}
 \ContractNumber{80NSSC21C0175}
 \Author{Filippov,Andrey; Filippova, Olga}
 \PerformingOrg{Elphel, Inc.\\
 1455 W 2200 S Ste 205 \\
 Salt Lake City, UT 84119-7214 \\}
 \SponsoringAgency{NASA \\
Shared Services Center (NSSC) \\
Building 1111, Jerry Hlass Road \\
Stennis Space Center MS 39529-0001 \\}
 \Abstract{{This report discusses the results of Phase~I effort to
prove the feasibility of dramatic improvement of the microbolometer-based LWIR
detectors sensitivity, especially for the 3D measurements. The resulting low
SWap-C thermal depth-sensing system will enable the situational awareness of
Autonomous Air Vehicles for Advanced Air Mobility. It will provide robust 3D
information of the surrounding environment, including low-contrast static and
moving objects, at far distances in degraded visual conditions and GPS-denied areas.
Our multi-sensor 3D perception enabled by COTS uncooled thermal sensors
mitigates major weakness of LWIR sensors - low contrast by increasing the system
sensitivity over an order of magnitude.
There were no available thermal image sets suitable for evaluating this
technology, making datasets acquisition our first goal.
We discuss the design and construction of the prototype system with sixteen
\SI{640x512}{\pixel} LWIR detectors, camera calibration to subpixel resolution,
capture, and process synchronized image. The results show the $3.84\times$
contrast increase for intrascene-only data and an additional $5.5\times$ – with
the interscene accumulation, reaching system noise-equivalent temperature
difference (NETD) of
\SI{1.9}{\milli\kelvin} with the \SI{40}{\milli\kelvin} sensors.}}

 \NumberPages{20}


\end{document}